\documentclass[final,1p,times]{elsarticle}

\usepackage{amssymb}
\usepackage{algorithm}
\usepackage{algpseudocode}
\usepackage{amsmath}
\usepackage{lscape}
\usepackage[labelfont=bf]{caption}
\usepackage{hyperref}
\usepackage{booktabs}
\usepackage{multirow}
\usepackage[toc,page]{appendix}
\usepackage{dirtytalk}
\usepackage{subcaption}
\usepackage{listings}
\usepackage[utf8]{inputenc}
\usepackage[T1]{fontenc}
\usepackage{graphicx} 

\makeatletter
\def\ps@pprintTitle{%
 \let\@oddhead\@empty
 \let\@evenhead\@empty
 \let\@oddfoot\@empty
 \let\@evenfoot\@oddfoot}
\makeatother

\begin{document}

\begin{frontmatter}

\title{Multi-refined Feature Enhanced Sentiment Analysis  Using 
Contextual Instruction}

\author[add1]{Peter Atandoh}
\author[add1]{Jie Zou $^*$\corref{cor1}}
\ead{jie.zou@uestc.edu.cn}
\author[add2]{Weikang Guo}
\author[add1]{Jiwei Wei}
\author[add3]{Zheng Wang}

\address[add1]{School of Computer Science and Engineering, University of Electronic Science and Technology of China, No. 2006, Xiyuan Avenue, Chengdu, 611731, China}
\address[add2]{Southwestern University of Finance and Economics, No. 555, Dongda Street, Chengdu, 611130, China}
\address[add3]{Tongji University, 1239 Siping Road, Shanghai, 200092, China}

\begin{abstract}
Sentiment analysis using deep learning and pre-trained language models (PLMs) has gained significant traction due to their ability to capture rich contextual representations. However, existing approaches often underperform in scenarios involving nuanced emotional cues, domain shifts, and imbalanced sentiment distributions.
We argue that these limitations stem from inadequate semantic grounding, poor generalization to diverse linguistic patterns, and biases toward dominant sentiment classes.
To overcome these challenges, we propose CISEA-MRFE, a novel PLM-based framework integrating Contextual Instruction (CI), Semantic Enhancement Augmentation (SEA), and Multi-Refined Feature Extraction (MRFE). CI injects domain-aware directives to guide sentiment disambiguation; SEA improves robustness through sentiment-consistent paraphrastic augmentation; and MRFE combines a Scale-Adaptive Depthwise Encoder (SADE) for multi-scale feature specialization with an Emotion Evaluator Context Encoder (EECE) for affect-aware sequence modeling.
Experimental results on four benchmark datasets demonstrate that CISEA-MRFE consistently outperforms strong baselines, achieving relative improvements in accuracy of up to 4.6\% on IMDb, 6.5\% on Yelp, 30.3\% on Twitter, and 4.1\% on Amazon. These results validate the effectiveness and generalization ability of our approach for sentiment classification across varied domains.
\end{abstract}

\begin{keyword}
Sentiment Analysis; Pre-trained Language Model; Multi-Refined Feature Extraction; Contextual Instruction

\end{keyword}

\end{frontmatter}

\section{Introduction}

Online evaluations have become pivotal in shaping user experience across domains such as e-commerce, media, and social platforms. These evaluations serve as rich sources of user sentiment and offer valuable insights into public perception. Sentiment Analysis (SA), a subfield of Natural Language Processing (NLP), applies computational techniques to extract and classify emotional tones from such data, enabling a deeper understanding of consumer preferences and opinions \citep{li2024sentiment, anitha2025sentiment, verma2025navigating}. For example, positive sentiment extracted from reviews can inform targeted marketing strategies and enhance recommendation systems.

Despite significant advances, SA remains a challenging task due to its reliance on nuanced linguistic cues, contextual semantics, and domain-specific expressions. Recent research has leveraged Pre-trained Language Models (PLMs) such as BERT \citep{devlin-etal-2019-bert}, RoBERTa \citep{DBLP:journals/corr/abs-1907-11692}, and GPT-2 \citep{radford2019language}, which are known for their ability to learn rich semantic representations from large corpora \citep{ghasemi2023deep, akdemir2024review, abdullah2022deep, wang2024incorporating}. Additionally, prompt-based learning methods \citep{dey2024aspect, liu2023gpt, ZHU2023103462, zhu2025soft, DBLP:conf/acl/XieAA24, gu2024agcvt} and hybrid deep learning architectures \citep{rasappan2024transforming, chauhan2023aspect, tao2025deep, jeon2024deep} have improved task-specific performance.

However, these existing models inadequately address the role of emotional nuance and context ambiguity in sentiment expressions, especially in longer or domain-specific reviews. Some studies have attempted to incorporate emotion constructs into SA \citep{DBLP:journals/tnn/HuangLYZZQ22,ghosh2023multitasking, DBLP:conf/wassa/Barnes23}, but often fall short in accurately identifying discrete emotional states or in achieving efficient feature extraction. Moreover, domain adaptation remains difficult, as PLMs trained on generic corpora may not effectively capture domain-specific sentiment cues \citep{ALAM2020129, JIA2021107423}.

To address these gaps, this study introduces a novel textual representation framework designed to extract specific emotions from reviews, thereby enhancing the understanding of human perceptions. Drawing inspiration from PLMs \citep{devlin-etal-2019-bert} and prompt-based pre-trained learning \citep{DBLP:journals/itc/ShiZSYLWW23, liu2023pre}, we design a Contextual Instruction (CI), a domain-aware contextual instruction mechanism that leverages PLMs to capture fine-grained emotional information. This module embeds relevant domain-specific context to improve the model's understanding of sentiment ambiguity and nuances. It supports the analysis of dynamic sentiments in long reviews and offers deeper insights into how specific topics, products, and services emotionally resonate with users. We also introduce instruction templates that adapt to domain-specific data, providing structured guidance to shape the model’s interpretation.
Furthermore, we introduce a Semantic Enhancement Augmentation (SEA) to enrich the dataset and improve generalization. To achieve this, we employ BERT-based semantic augmentation using the T5 model to paraphrase input text while preserving meaning. This exposes the model to diverse sentence structures and strengthens its semantic robustness. Additionally, we design a Multi-Refined Feature Extraction (MRFE), a context-aware BERT method for sentiment-indicative (Local) patterns and global feature extraction, named BERT-MRFE. Our approach utilizes emotional cues to extract complex features from textual reviews, enhancing classification performance. Specifically, we employ Scale-Adaptive Depth-Wise Encoder (SADE) and an Emotional Evaluator Context Encoder (EECE) based on the BiLSTM architecture to enrich BERT representations. Compared to prior approaches, our SADE strategy significantly improves locality-aware representation and granularity of feature extraction by leveraging multi-scale depth-wise separable convolutions, while preserving computational efficiency through channel-wise filtering and grouped processing. This design aligns with recent findings emphasizing the role of receptive field diversity and efficient convolutional decomposition for text understanding \citep{kaiser2017depthwise, wu2019pay, howard2017mobilenets}.
Crucially, unlike prior work that utilizes only pooled classification vectors \citep{zhu2023bert}, our EECE incorporates a gating mechanism and attention weights to emphasize emotionally salient features and regulate information flow. This yields a more expressive classification vector that highlights the significance of multi-refined feature extraction in SA.
In summary, our proposed model analyzes input textual reviews across various domains to extract emotional signals and salient textual features. By integrating CI, BERT, SEA, and MRFE, the model effectively combines contextual semantics with enriched instruction-guided representations. This architecture enables the modeling of interactions between word combinations while incorporating both sentiment and emotional information. Through the use of contextual semantic instructions and multi-refined features, this study delivers a comprehensive and robust framework for sentiment analysis.

The main contributions of this work are as follows\footnote{We will make our code publicly available upon acceptance}:
\begin{itemize}
    \item We propose a novel sentiment analysis framework, CISEA-MRFE, that integrates contextual instruction and semantic augmentation to enhance emotional feature extraction in PLM-based architectures.
    \item We design instruction templates that are adaptable to domain-specific datasets, enabling the transformation of review sentences into enriched contextual forms while preserving sentiment polarity.
    \item We introduce a Multi-Refined Feature Extraction (MRFE) module that combines a Scale-Adaptive Depth-wise Encoder with an Emotional Evaluator Context Encoder to capture both local and global emotional representations within text.
    \item We conduct extensive evaluations on four benchmark sentiment datasets, demonstrating that CISEA-MRFE achieves superior performance in accuracy and macro-F1 score compared to competitive baseline models.
\end{itemize}

The remainder of this paper is organized as follows: Section 2 reviews related work on sentiment analysis, instruction-based learning, and emotion-aware modeling. Section 3 presents the proposed CISEA-MRFE framework and its components. Section 4 describes the experimental setup and reports comparative results. Section 5 concludes with key findings and directions for future research.

\section{Related Work}

\subsection{Sentiment Classification}
Sentiment classification can be broadly categorized into lexicon-based, classical machine learning, and deep learning approaches \citep{anitha2025sentiment}. Lexicon-based sentiment SA involves classifying text as positive, negative, or neutral based on pre-defined dictionaries. This technique typically builds classifiers from tagged text or phrases. To accurately capture the emotional tone of negated expressions, specialized sentiment lexicons are necessary \citep{DBLP:journals/jair/KiritchenkoZM14, SHANG2023103187}. For example, some studies propose automatic generation of sentiment dictionaries tailored to specific objectives \citep{DBLP:journals/eswa/WuWCWH19}. SENTIWORDNET is one such lexical resource that assigns objective, positive, and negative scores to each WORDNET synset \citep{Esuli2006SENTIWORDNETAP}. While lexicon-based methods are simple and interpretable, they suffer from limitations due to variations in language, context, and domain, and are generally ineffective at handling sarcasm or humor.

Classical machine learning algorithms such as Support Vector Machines (SVMs), Decision Trees, Naive Bayes (NB), and Maximum Entropy have also been applied to SA tasks \citep{10007390}. For instance, the Elman Neural Network, optimized via the Local Search Improvised Bat algorithm, has been used for textual sentiment classification \citep{ZHAO2021102656}. Other methods include fuzzy cognitive maps for emotion categorization \citep{JAIN2022102758} and comparative studies evaluating SVM and NB performance \citep{PAVITHA2022279}. While classical methods benefit from efficient training and good baseline performance with feature engineering, they often struggle with multi-class classification, high-dimensional data, and complex non-linear relationships.

Deep learning has significantly advanced sentiment classification, achieving high performance \citep{JAIN2022102758}. Models such as feedforward neural networks (FNNs) and recurrent neural networks (RNNs) are widely used. These models often utilize sophisticated word representations such as Word2Vec \citep{JIA2021107423}, GloVe \citep{kumar2024aspect}, FastText \citep{KHASANAH2021343}, and BERT \citep{devlin-etal-2019-bert}. Alam et al. \cite{ALAM2020129} developed domain-specific word embeddings from social media data, integrating them into a D-CNN with global average pooling. Ling et al. \cite{9108541} proposed multichannel CNNs with varying filter widths, while Gan et al. \citep{GAN2020116} introduced a hierarchical CNN with self-attention for multi-entity sentiment tasks. 
Bin et al. proposed Sentic GCN, an aspect-based sentiment analysis model that integrates graph convolutional networks with SenticNet-derived affective knowledge to explicitly model feature opinion dependency structures \citep{liang2022aspect}. 
Atandoh et al. \cite{9674171} combined deep CNN with BiLSTM and embedding layers, and Pimpalkar et al. \cite{PIMPALKAR2022117581} integrated CNN with MBiLSTM for feature extraction. Liao et al. \citep{LIAO2022102934} proposed a sentiment knowledge graph with orthogonal attention, while Jia et al. \citep{JIA2022108032} combined BERT, CNN, and attention mechanisms for improved performance. These deep learning methods have shown promise across multiple modalities, but often lack reasoning capabilities. Prompt-based learning \citep{DBLP:journals/corr/abs-2205-07220} has emerged as a key enhancement, enabling structured inputs that better capture sentiment nuances. 

Despite notable advancements, existing studies often overlook key challenges this paper addresses. Specifically, the integration of localized emotional representation with deep PLM-based contextual reasoning, especially when guided by structured task instructions, remains underexplored.
Also, few works systematically evaluate combined multi-scale semantic extraction and affective encoding across diverse domains or PLM architectures.
Furthermore, instruction-driven sentiment models typically focus on high-level classification or mask-filling, and rarely examine how instruction frameworks interact with emotion-centric modules like ours.
These gaps motivate our CISEA-MRFE architecture, which explicitly combines CI, SEA, and MRFE to tackle multi-granular semantic reasoning and emotion awareness in PLM-based sentiment analysis.

\subsection{Emotion in Sentiment Analysis}

Emotion plays a pivotal role in advancing sentiment classification beyond simplistic polarity labels. Traditional sentiment models often reduce textual inputs to binary or ternary sentiment outcomes, overlooking the nuanced emotional undercurrents embedded in natural language. To address this, recent literature has introduced various mechanisms for incorporating emotional signals. 

To begin with, emotion representation has evolved from static lexicon-based methods \citep{salovey1990emotional} to more adaptive, neural-based embeddings that capture affective semantics \citep{kumar2022discovering}. For instance, models using emotion lexicons like NRC or WordNet-Affect embed emotional tags directly into the input space but lack context sensitivity. In contrast, recent studies by Rameezunnisa et al and Huang et al. \citep{RAMEEZUNNISA202179, DBLP:journals/tnn/HuangLYZZQ22} propose attention-enhanced and psychologically grounded embeddings that dynamically adapt to context. Similarly, Zhihan et al. \citep{zhang2024cogaware} introduced CogAware, a cognitively inspired model that jointly learns textual and cognitive feature representations using multi-task optimization, enabling purified domain-specific and cross-domain embeddings for robust, emotion-aware sentiment analysis across heterogeneous data sources .

Emotion integration techniques also differ significantly. Some works utilize attention mechanisms to focus on emotionally salient tokens, while others employ gating strategies or fusion networks to merge emotional and semantic features \citep{DBLP:journals/ipm/LiangWZ24}. However, many of these methods do not address the hierarchical nature of emotional expression, e.g., token-level emotions vs. sentence-level tone. Furthermore, emotion granularity and generalizability remain underexplored. Most models rely on coarse-grained emotion labels, ignoring the multidimensional nature of emotions \citep{wankhade2022survey, buechel2018emotion}. Wankhade et al. suggest that most emotion recognition models classify emotions into a few basic categories (e.g., joy, anger, sadness) and rarely consider richer formulations like Plutchik’s wheel or dimensional spaces such as valence-arousal \cite{wankhade2022survey}. Buechel et al. also discuss the limitations of discrete label-based approaches and promote valence-arousal-dominance models as more expressive alternatives \cite{buechel2018emotion}. Additionally, emotional representations often fail to generalize across domains due to cultural variance or source-dependent cues (e.g., sarcasm on Twitter vs. product reviews).

Despite these innovations, several research gaps remain. First, emotion is rarely modeled as a dynamic, multi-scale, and context-aware construct. Second, prior models often lack bidirectional dependency propagation mechanisms and fail to unify emotional modeling with semantic granularity or instruction-aware disambiguation. Third, many systems ignore negation, implicit sentiment, or emotional inversion challenges that are central to real-world interpretability.

To overcome these limitations, we reconceptualize emotion as a dynamic, multi-scale, and context-aware signal. Our proposed MRFE module incorporates a sequence encoder-based Emotional Evaluator Context Encoder regulated through attention and emotional gating mechanisms. Unlike static approaches, this allows the model to dynamically adjust to latent affective cues, manage emotional overlap, and distinguish between implicit and explicit sentiment expressions. This design enables robust, fine-grained emotion modeling that improves generalization across tasks and domains, effectively addressing prior limitations in emotional signal integration.

\subsection{Instructing and Prompting}

In PLMs, instruction or prompting refers to embedding task-related information into the input to guide model behavior \citep{schick2020exploiting, gao2020making}. Prompt-based learning enables a single model to perform multiple tasks without retraining \citep{shin2020autoprompt}. Liu et al. \cite{liu2023pre} proposed a unified framework integrating pre-training, prompting, and prediction. Prompt learning typically involves three stages: prompt insertion, response generation, and response mapping \citep{liu2023pre}. Prompts are inserted using templates with masked or empty response slots \citep{petroni2019language, DBLP:conf/emnlp/LesterAC21}. Response mapping links generated outputs to target labels, aligning the prompt input with the pre-trained task objective.

Several studies have explored prompt-based sentiment analysis. Sun et al. \citep{sun2019utilizing} introduced a method for quickly generating auxiliary sentences. Liu et al. \cite{liu2023pre} proposed a simple prompt framework for SA, while Song et al. \citep{DBLP:journals/corr/abs-1902-09314} developed BERT-SPC, a sentence-pair prompt model. More recently, Xinjie et al. introduced a Prompt Tuning with Domain Knowledge  model that integrates domain knowledge into the pre-training framework via prompt tuning, improving domain-specific sentiment understanding \citep{sun2024harnessing}.

However, key gaps remain. First, existing work rarely explores how instructional conditioning interacts with multi-level semantic reasoning and emotional evaluation. Few models have systematically examined instruction-enhanced sentiment analysis in conjunction with refined feature encoders. Second, LLMs continue to struggle with subtle sentiments and ambiguous contextual cues. Third, instruction-based prompting has largely been confined to few-shot template tuning or zero-shot reasoning approaches that fall short under domain shifts or when nuanced disambiguation is required.

CI addresses these challenges by embedding relevant background information e.g., appending “This is a movie review” to help clarify ambiguous expressions like “It was magnificent.” Incorporating CI into the CISEA-MRFE pipeline enhances model accuracy, interpretability, and generalization across diverse domains, providing a robust solution for real-world sentiment analysis tasks.
\begin{figure}[htbp]
    \centering
    \includegraphics[width=0.95\linewidth]{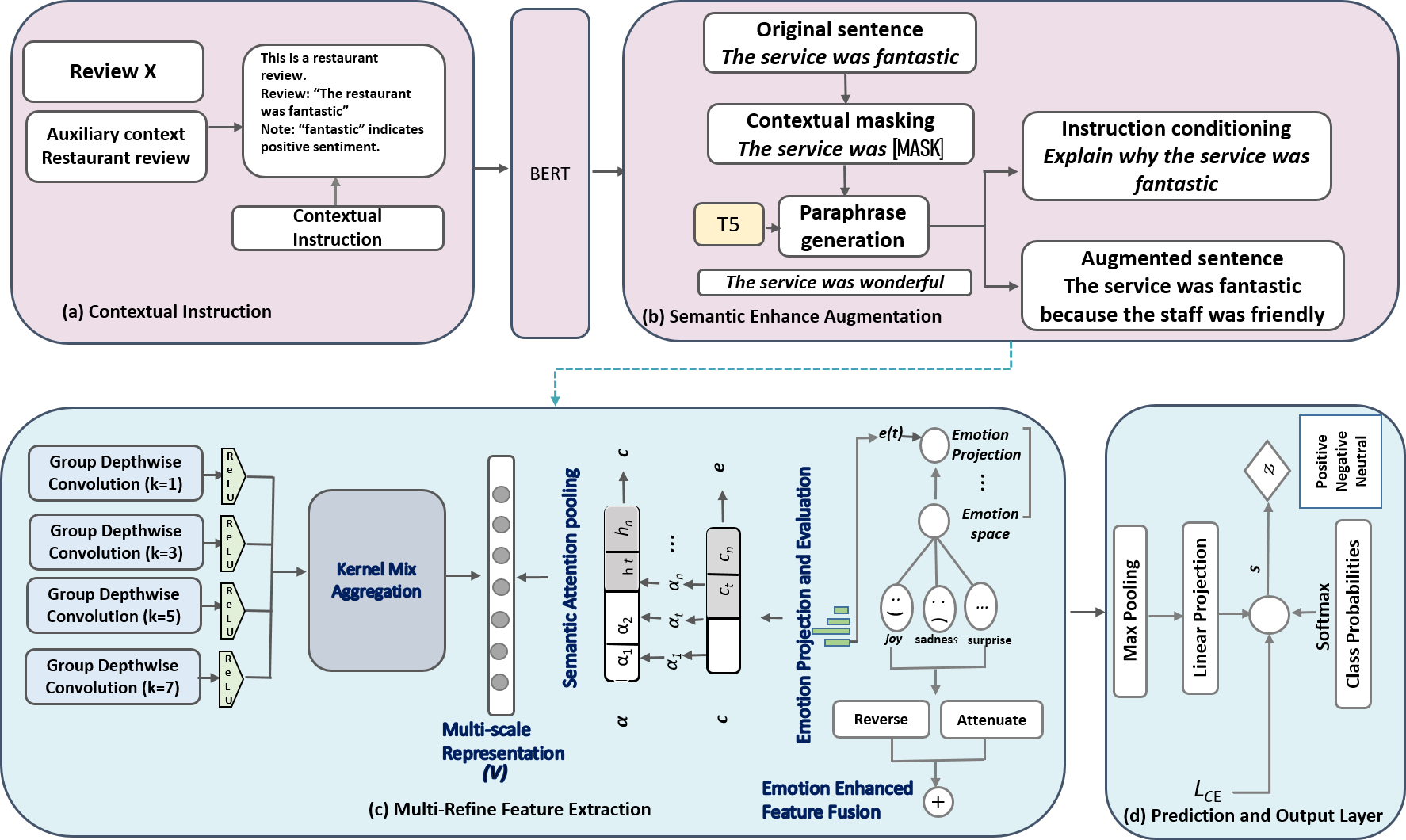}
    \caption{Overall framework of the proposed CISEA-MRFE model.}
    \label{Figure_1}
\end{figure}

\section{Methodology}   
Figure \ref{Figure_1} depicts the structure of our proposed model CISEA-MRFE.  
The CISEA-MRFE can be divided into four main components: (1) the CI instructs the input text and guides BERT to improve the extraction of pertinent information and emotion as well as context-aware token embeddings; 
(2) a SEA further transforms the text reviews to contextually similar instances while retaining their meaning; (3) an MRFE layer captures local and global sequential features and extensive contextual information; and (4) a predictive and output layer. The output is subsequently classified as positive, negative, or neutral by the softmax layer. 

A notable distinction in our methodology lies in using contextual instruction to dynamically adjust the instruction template based on the nature of the input text, thereby improving BERT's focus on extracting rich embeddings and emotions. The semantic enhancement augmentation also improves textual ambiguity for the model to comprehend. To further enhance feature extraction from text, we use a multi-refined feature extractor that captures sentiment inductive patterns and global semantic features. This unique architectural design allows our model to effectively understand textual language, enhanced by Contextual Instruction tailored to the specific context of each input combined with a Multi-Refined Feature Extractor for handling local and global sequential patterns. 

\subsection{Problem Definition}

We define sentiment classification as predicting a polarity label
\(
y \in \mathcal{S} = \{\text{positive},\ \text{negative}, \newline \ \text{neutral}\}
\)
for each review \(r = (x_1, \dots, x_m)\), where each sentence
\(x_i = (w_{i,1}, \dots, w_{i,n_i})\), and \(w_{i,j} \in \mathcal{V}\). 
The review is first enriched with contextual instruction:
\(
A = \mathrm{CI}(r) \in \mathbb{R}^{n \times d_w}
\)
where \(n\) is the token length after augmentation and \(d_w\) is the embedding dimension.
Our model maps this to sentiment probabilities:
\(
\hat{\mathbf{y}} = f_{\theta}(A) = \mathrm{CISEA\text{-}MRFE}(A)
\in \Delta^{|\mathcal{S}|}, \quad
\hat{y} = \arg\max_s \hat{y}_s
\)
where \(\Delta^{|\mathcal{S}|}\) is the probability simplex.

\subsection{Context Instruction}

Let the original review be \(X = (w_1, w_2, \dots, w_n)\), and let the optional auxiliary context \(C\) include domain tags or prior sentences (e.g., “restaurant review”). We apply the CI function to obtain an instruction-augmented input:

\begin{equation}
X_{\mathrm{instr}} = \mathrm{CI}(X, C),
\end{equation}
where \(\mathrm{CI}(\cdot)\) prepends a structured directive (e.g., “This is a restaurant review.”) and may optionally highlight key emotion-bearing tokens to guide the encoder’s attention.

When \(X_{\mathrm{instr}}\) is encoded by the BERT-based model, it produces a probability distribution over sentiment classes:

\begin{equation}
\hat{\mathbf{y}} = P\big([S] \mid X_{\mathrm{instr}}; \theta\big), \quad \hat{\mathbf{y}} \in \Delta^3
\end{equation}
where \(\Delta^3\) is the 3-dimensional probability simplex corresponding to sentiment classes \(\{\text{positive},\newline,\text{negative}, \text{neutral}\}\). 

This formulation emphasizes how CI augments the input to steer the encoder toward sentiment-relevant cues and domain context without performing a hard classification decision at this stage.

\subsubsection{Instruction Template Example}
Below is an instantiation of our template, demonstrating how raw review text is transformed into an instruction-rich input:

\begin{lstlisting}[basicstyle=\ttfamily\small,frame=single]
Review: "The restaurant was fantastic."
Comment: The review is positive because  "fantastic" expresses 
          strong sentiment.
\end{lstlisting}

In this example, the prefix “Review:” clearly signals the input segment. The phrase “Comment: The review is positive because …” acts as a directive that reinforces task intent and explicitly highlights the sentiment-laden token. Emotion-relevant keyword “fantastic” is directly mentioned, strategically guiding the model to attend to critical evidence.

The use of explicit connectors (“because …”) encourages the model not merely to predict sentiment but also to establish a rationale a mechanism that has been shown to boost interpretability and robustness in instruction-tuned language models.

\subsubsection{Formal Template Construction}  
We define the instruction template as:
\begin{equation}
T = f(X, R, S, C, I; P),
\end{equation}
where 
\(R\) is the label response (e.g., “The review is positive.”),  
\(I\) is the sentiment identifier (e.g., “fantastic”), and \(P\) is the chosen template pattern.

The constructed instruction-augmented sentence is then expressed as:
\begin{equation}
X' = T(X, R, S, C, I; P),
\end{equation}
where \(X'\) denotes the enriched text obtained by fusing the original input with the contextual and sentiment-aware components.

Finally, the augmented sentence \(X'\) is tokenized as:
\begin{equation}
A = \mathrm{Tokenizer}(X') \in \mathbb{R}^{n \times d_w},
\end{equation}
where \(n\) is the token count and \(d_w\) is the embedding dimension.

\subsubsection{Template Generation Strategy}  
Templates are manually crafted for clarity and reliability~\citep{schick2020exploiting}. Each review \(X\) is paired with its corresponding template via:
\begin{equation}
\tilde{P} = C + X.
\end{equation}
This ensures that BERT receives explicit contextual and sentiment guidance.
By injecting both semantics (e.g., “This is a restaurant review.”) and affective cues (e.g., “fantastic”), the CI module primes the model to better attend to sentiment-relevant patterns before deeper MRFE processing.

\subsection{BERT Encoding Layer}

Bidirectional Encoder Representations from Transformers (BERT) is highly effective for modeling sentiment across long textual sequences due to its deep bidirectional attention, multi-layer transformer architecture, and contextualized representations. In this work, we employ a pre-trained BERT model to encode the instruction-augmented input text.

Let the instruction-augmented input be denoted as \( A \in \mathbb{R}^{n \times d_w} \), where \( n \) is the number of tokens in the sequence and \( d_w \) is the token embedding dimension. This input is passed through the BERT encoder to generate deep contextual embeddings:
\begin{equation}
    \mathbf{E} = \text{BERT}(A) = [\mathbf{e}_1, \mathbf{e}_2, \ldots, \mathbf{e}_n], \quad \mathbf{E} \in \mathbb{R}^{n \times d},
\end{equation}
where \( \mathbf{e}_i \in \mathbb{R}^{d} \) represents the contextual embedding of the \( i \)-th token, and \( d = 768 \) for BERT-base.

These embeddings incorporate both semantic and syntactic context, guided by the instruction-enhanced input. The instruction prefix helps BERT attend to sentiment-bearing phrases, emotion indicators, and domain-specific cues in the review. This embedding matrix \( \mathbf{E} \) serves as input to subsequent layers such as the Multi-Refined Feature Extractor (MRFE) for deeper semantic and emotional reasoning.

\subsection{Semantic Enhancement Augmentation}

Data augmentation enhances model robustness by increasing training sample diversity. In sentiment analysis, paraphrastic variation is particularly valuable for generalization across domains and linguistic styles.

We propose SEA, leveraging T5-based paraphrasing to enrich the dataset with sentiment-preserving text. Formally, for each original sentence \(x\), we generate augmented variants:
\begin{equation}
x_k' \sim \mathrm{T5}_{\text{paraphrase}}(x),\quad \text{s.t. } \mathrm{sentiment}(x_k') = \mathrm{sentiment}(x).
\end{equation}

The SEA pipeline includes:
\begin{enumerate}
  \item \textbf{Contextual masking}: Replace adjectives/nouns in \(x\) with \texttt{[MASK]}.
  \item \textbf{Paraphrase generation}: Use T5 to predict replacements, producing top-\(k\) candidates.
  \item \textbf{Instruction conditioning}: Apply contextual templates (see Sec.~3.2) to guide the paraphrase toward stylistic or emotional clarity.
  \item \textbf{Style variation and explanation}: Generate formal/informal variants and optionally append emotion rationale, e.g.,: 
\end{enumerate}
\begin{lstlisting}[basicstyle=\ttfamily\small,frame=single]
The service was fantastic because the  staff was friendly.
  \end{lstlisting}
To scale efficiently, we employ batch-wise masking and generation. All augmented samples are validated for sentiment consistency via a classifier-filter loop. Empirically, SEA increases semantic variance while retaining polarity integrity, aligning with best practices in prompting-based data augmentation \citep{DBLP:conf/acl/0003XSHTGJ22}.

\subsection{Multi-Refine Feature Extraction (MRFE)}

The MRFE module is the third architectural component following the CI and SEA layers. It is designed to extract both localized and global affective-semantic representations. MRFE comprises two primary subcomponents: (1) SADE, to capture hierarchical n-gram patterns via channel-wise depthwise convolutions, and (2) EECE, which models long-range dependencies while modulating representations with emotion-sensitive attention and gating mechanisms.

\subsubsection{Scale-Adaptive Depthwise Encoder (SADE)}

To extract sentiment-relevant features across varying n-gram scales, the SADE module applies grouped depth-wise convolutions with multiple kernel sizes to capture diverse local patterns. Let the input be \( \mathbf{E} \in \mathbb{R}^{n \times d} \), the contextualized embeddings from the previous layer. The SADE operation is defined as:
\begin{equation}
    \mathbf{V} = \frac{1}{K} \sum_{k \in \mathcal{K}} \text{ReLU}\left(\text{Conv}^{(k)}_{\mathrm{dw}}(\mathbf{E})\right), \quad \mathbf{V} \in \mathbb{R}^{n \times c}
\end{equation}
where \( \mathcal{K} = \{1, 3, 5, 7\} \), which is a set of kernel sizes defining multi-scale receptive fields. \( \text{Conv}^{(k)}_{\mathrm{dw}}: \mathbb{R}^{n \times d} \rightarrow \mathbb{R}^{n \times c} \), indicating 1D grouped depth-wise convolution with kernel size \( k \) and output channel dimension \( c \). \( K = |\mathcal{K}| = 4 \), which is the total number of convolutional branches.

Each branch applies ReLU activation after the depth-wise convolution to introduce non-linearity. The resulting feature maps are averaged across kernel sizes to produce the final representation \( \mathbf{V} \in \mathbb{R}^{n \times c} \), which captures fine-to-coarse contextual information.

This architectural design enables localized feature specialization by allowing each feature channel to learn independently while promoting parameter efficiency through grouped operations. Inspired by prior works in convolutional text classification \citep{zhang2015character, yin2017comparative}, this configuration enhances the model's ability to learn multi-scale sentiment cues that differ across domains.
It is crucial to note that the scale-adaptive receptive fields improve semantic granularity in sentiment modeling while maintaining computational efficiency due to the low-overhead depth-wise convolutional design \citep{howard2017mobilenets, sandler2018mobilenetv2}.

\subsubsection{Emotion Evaluator Context Encoder (EECE)}

The EECE module refines multi-scale features \( \mathbf{V} \in \mathbb{R}^{n \times c} \) by modeling sequential context and integrating emotion-based modulation.

\paragraph{Contextual Sequence Modeling}  
A contextual sequence encodes forward and backward context:
\begin{equation}
\mathbf{H} = \text{BiLSTM}(\mathbf{V}) = [\mathbf{h}_1, \ldots, \mathbf{h}_n], \quad \mathbf{H} \in \mathbb{R}^{n \times 2h},
\end{equation}
where each \(\mathbf{h}_t = [\overrightarrow{\mathbf{h}}_t; \overleftarrow{\mathbf{h}}_t] \in \mathbb{R}^{2h}\), and \( h \) is the hidden-state dimension.

\paragraph{Semantic Attention Pooling}  
We apply additive attention to summarize context:
\begin{equation}
\mathbf{u}_t = \tanh(\mathbf{W}_a \mathbf{h}_t + \mathbf{b}_a), \quad \alpha_t = \frac{\exp(\mathbf{v}_a^\top \mathbf{u}_t)}{\sum_j \exp(\mathbf{v}_a^\top \mathbf{u}_j)}, \quad \mathbf{c} = \sum_t \alpha_t \mathbf{h}_t,
\end{equation}
with learnable parameters \(\mathbf{W}_a \in \mathbb{R}^{d_a \times 2h}\), \(\mathbf{b}_a \in \mathbb{R}^{d_a}\), and \(\mathbf{v}_a \in \mathbb{R}^{d_a}\).

\paragraph{Emotion Projection \& Evaluation}  
Each \(\mathbf{h}_t\) is projected to emotion-space:
\begin{equation}
\mathbf{e}_t = \mathrm{softmax}(\mathbf{W}_e \mathbf{h}_t + \mathbf{b}_e), \quad \mathbf{e}_t \in \Delta^{|E|}, 
\end{equation}
with \(\mathbf{W}_e \in \mathbb{R}^{|E| \times 2h}\), \(\mathbf{b}_e \in \mathbb{R}^{|E|}\).  
Emotion compatibility scores are computed as:
\begin{equation}
Y_e(w_t) = \frac{\mathbf{E}_{w_t}^\top \mathbf{p}_e}{\|\mathbf{E}_{w_t}\|\|\mathbf{p}_e\|}, \quad e = 1,\dots,|E|,
\end{equation}
where \(\mathbf{E}_{w_t}, \mathbf{p}_e \in \mathbb{R}^d\).  
Negation-aware modulation is applied per:
\begin{equation}
Y_e(w_t) \leftarrow
\begin{cases}
-\,Y_e(w_t), & \text{if reversed},\\
\lambda \cdot Y_e(w_t), & \text{if attenuated},\ \lambda\in(0,1).
\end{cases}
\end{equation}

\paragraph{Emotion-Enhanced Feature Fusion}  
Emotion-weighted features are fused residually:
\begin{equation}
\tilde{\mathbf{h}}_t = \mathbf{h}_t + \sum_{e=1}^{|E|} \alpha_e\,Y_e(w_t), \quad \tilde{\mathbf{c}} = \sum_{t=1}^n \alpha_t\,\tilde{\mathbf{h}}_t,\quad \tilde{\mathbf{c}} \in \mathbb{R}^{2h},
\end{equation}
where \(\alpha_e\in\Delta^{|E|}\) are emotion attention weights and \(\alpha_t\) are semantic attention scores from earlier.  
This residue-based fusion ensures that emotion cues refine rather than override contextual semantics \citep{DBLP:journals/tnn/HuangLYZZQ22}.

\subsection{Prediction and Output Layer}

Given the sequence of emotion-aware hidden states from the EECE module, denoted as \( \tilde{\mathbf{H}} = [\tilde{\mathbf{h}}_1, \tilde{\mathbf{h}}_2, \ldots, \tilde{\mathbf{h}}_n] \in \mathbb{R}^{n \times 2h} \), we apply a max pooling operation along the temporal axis to obtain a fixed-length representation:
\begin{equation}
    \mathbf{h}_{\text{final}} = \max(\tilde{\mathbf{H}}, \text{dim}=1), \quad \mathbf{h}_{\text{final}} \in \mathbb{R}^{2h}. 
\end{equation}
This max-pooled vector summarizes the most salient features across time, capturing the strongest emotional and contextual signals.
 
The pooled vector \( \mathbf{h}_{\text{final}} \) is fed into a fully connected linear projection layer followed by a softmax activation function to compute class probabilities:
\begin{equation}
    \hat{\mathbf{y}} = \text{softmax}(\mathbf{W}_o \mathbf{h}_{\text{final}} + \mathbf{b}_o), \quad \hat{\mathbf{y}} \in \mathbb{R}^{C}.
\end{equation}
Here, \( \mathbf{W}_o \in \mathbb{R}^{C \times 2h} \) and \( \mathbf{b}_o \in \mathbb{R}^{C} \) are learnable output projection parameters, and \( C \) is the number of sentiment classes (e.g., positive, negative, neutral).
 
The model is trained using the cross-entropy loss:
\begin{equation}
    \mathcal{L}_{\text{CE}} = -\sum_{c=1}^{C} y_c \log(\hat{y}_c), \quad y_c \in \{0,1\}, \quad \sum_{c=1}^{C} y_c = 1.
\end{equation}
\( \mathbf{y} = [y_1, \ldots, y_C] \) is the one-hot encoded ground-truth label vector and \( \hat{\mathbf{y}} \) is the predicted class distribution.

At inference time, the predicted class \( \hat{z} \) is determined via the maximum a posteriori probability:
\begin{equation}
    \hat{z} = \arg\max_{c \in \{1, \ldots, C\}} \hat{y}_c.
\end{equation}

This prediction layer leverages the emotionally and contextually enriched representation \( \mathbf{h}_{\text{final}} \), enabling robust and interpretable sentiment classification across diverse textual domains.

\section{Experiment and Analysis} 

To validate the effectiveness of our proposed CISEA-MRFE model, we aim to answer the following research questions:

\begin{itemize}
\item \textbf{RQ1}: How competitive is the proposed model compared to current state-of-the-art baselines? 
\item \textbf{RQ2}: To what extent do the components of CISEA-MRFE enhance sentiment classification by improving cue identification, mitigating polarity imbalance, and reducing performance variance?
\item \textbf{RQ3} How do architectural modules influence computational efficiency and scalability in sentiment classification across heterogeneous datasets?
\item \textbf{RQ4}: How do the different parameters within the proposed model affect its overall performance?
\end{itemize}
 
\subsection{Experimental Setup}

\subsubsection{Datasets}
We utilize four real-world benchmark datasets to evaluate the performance of our CISEA-MRFE model. Table \ref{tab:my_label_1} summarizes their statistics, including label distributions and average sentence lengths. All datasets are widely used in the sentiment analysis literature \citep{DBLP:journals/tnn/HuangLYZZQ22,tang-etal-2016-effective, PIMPALKAR2022117581, ATANDOH2023101578}.

\textbf{IMDb}\footnote{\url{https://www.kaggle.com/datasets/lakshmi25npathi/imdb-dataset-of-50k-movie-reviews}}: This binary sentiment classification dataset contains 50,000 movie reviews and serves as a standard benchmark for NLP tasks \citep{yang2019xlnet, heinsen2022algorithm}.

\textbf{Yelp2014}\footnote{\url{https://www.kaggle.com/datasets/ilhamfp31/yelp-review-dataset/data}}: Comprising five sentiment labels ranging from 1 (very negative) to 5 (very positive), we randomly sample 50,000 reviews.

\textbf{Twitter}\footnote{\url{https://huggingface.co/datasets/carblacac/twitter-sentiment-analysis}}: This dataset includes binary sentiment labels (1 = positive, 0 = negative) and serves as a benchmark for social media sentiment classification \citep{10.1145/2872427.2883037}.

\textbf{Amazon}\footnote{\url{https://datarepo.eng.ucsd.edu/mcauley_group/data/amazon_2023/raw/review_categories/Electronics.jsonl.gz}}: Focused on the electronics domain, this dataset labels reviews with ratings below three stars as negative and above three stars as positive \citep{10.5555/2969239.2969312, 10.1145/3357384.3357973}. 

Following \cite{DBLP:journals/tnn/HuangLYZZQ22, yang2019xlnet}, all datasets, 80\% of instances are use for training, 10\% for validation, and 10\% for testing. 
For all datasets, we apply instruction templates as described in the contextual instruction and semantic enhancement augmentation sections. This expansion ensures balanced representation across sentiment polarities, thereby improving model training and generalization.

\subsubsection{Evaluation Metrics}
To assess model performance, we adopt standard metrics including accuracy and macro-F1 score, which are robust for both balanced and imbalanced datasets \citep{DBLP:journals/tnn/HuangLYZZQ22, DBLP:journals/corr/abs-1902-09314}:

\begin{equation}
    \text{Accuracy} = \frac{CP}{N},
\end{equation}
\begin{equation}
    \text{M-F1} = \frac{1}{N} \sum_{n=1}^{N} F1_i,
\end{equation}
where \(CP\) is the number of correct predictions, \(N\) is the total number of samples, and \(F1_i\) is the F1 score for class \(i\). The macro-F1 score computes the unweighted mean F1 across all classes, treating each class equally.

\subsubsection{Parameter Settings}

We employ L2 regularization with a weight decay coefficient of \(1 \times 10^{-5}\) and set the dropout rate to 0.1. The model is trained for 20 epochs using a batch size of 64 and a learning rate of \(2 \times 10^{-5}\).
To enhance adaptability across datasets, we perform a dynamic hyperparameter search specifically on the MRFE module. This includes evaluating maximum sequence lengths in \([32, 64, 128, 256, 512]\) and exploring fusion mechanisms namely attention-augmented feature stacking and summation-based fusion.
For scale-adaptive convolution, we set \(\texttt{groups} = 768\), implementing channel-wise depth-wise convolutions. This ensures each feature channel is processed independently, promoting localized feature specialization while significantly reducing computational overhead relative to standard convolutions.

Furthermore, we assess the impact of various kernel size combinations in the depth-wise CNN component, as detailed in Table~\ref{tab:2}. A sensitivity analysis is conducted to examine the configuration effects across datasets (see Section~5.2.6), with accuracy trends visualized in Figure~\ref{fig:5}. Full hyperparameter configurations are provided in Table~\ref{tab:8}.

\paragraph{Model Configuration of CI Variants} To systematically evaluate the contributions of different architectural components, we construct four variants of our contextual instruction-based framework by selectively integrating SEA and submodules of the MRFE layer:

\begin{itemize}
    \item \textbf{CISEA-SADE}: Integrates SEA and the SADE only, focusing on local syntactic pattern extraction through multi-kernel depthwise convolutions.
    
    \item \textbf{CISEA-EECE}: Integrates SEA and the EECE only, capturing global context and affective signals via bidirectional recurrent encoding and attention mechanisms.
    
    \item \textbf{CI-MRFE}: Excludes SEA, but retains the full MRFE module composed of SADE and EECE. This version emphasizes hierarchical feature refinement without semantic augmentation.
    
    \item \textbf{CISEA-MRFE (Full Model)}: Incorporates CI, SEA, SADE, and EECE. This complete variant jointly models semantic, local, global, and emotional dependencies for robust sentiment classification.
\end{itemize}

\begin{table}[htbp]
\centering
\small
\caption{Statistics of the benchmark datasets used in this study.}
\begin{tabular}{lcccc}
\toprule
\textbf{Dataset} & \textbf{Total Samples} & \textbf{Avg. Words/Sentence} & \textbf{Classes} \\
\midrule
IMDb     & 50,000   & 231 / 5  & 2 \\
Yelp     & 50,000   & 134 / 4  & 5 \\
Twitter  & 127,463  & 19 / 5   & 2 \\
Amazon   & 50,000   & 80 / 5   & 2 \\
\bottomrule
\end{tabular}
\label{tab:my_label_1}
\end{table}

\begin{table}[h]
\centering
\caption{
Configurations of Depth-Wise Convolutions with Varying Kernel Size Combinations. 
Each configuration employs groups=768, ensuring channel-wise independence in convolution.
The window size denotes the largest kernel used in the multi-scale feature extractor (SADE).
}
\begin{tabular}{lcc}
\hline
Kernel Sizes ($\mathcal{K}$) & Window Size (Max $k$) & Groups \\
\hline
$[1, 3, 5]$   & 5 & 768 \\
$[1, 3, 7]$   & 7 & 768 \\
$[3, 5, 7]$   & 7 & 768 \\
$[3]$         & 3 & 768 \\
$[5]$         & 5 & 768 \\
$[1, 3, 5, 7]$ & 7 & 768 \\
\hline
\end{tabular}
\label{tab:2}
\end{table}
\subsection{Full hyperparameter configurations}
Model configuration and training setup for CISEA-MRFE are shown in Table~\ref{tab:8}.

\begin{table}[htbp]
\centering
\small
\caption{Model configuration and training setup for CISEA-MRFE.}
\begin{tabular}{ll}
\toprule
\textbf{Component} & \textbf{Value} \\
\midrule
Pretrained Language Model & \texttt{bert-base-uncased} \\
Tokenizer & \texttt{BertTokenizer} \\
Maximum Sequence Length & 512 \\
Batch Size & 64 \\
Epochs & 20 \\
Optimizer & AdamW \\
Learning Rate & $2 \times 10^{-5}$ \\
Loss Function & Cross-Entropy Loss \\
Depthwise CNN Kernel Sizes & [1, 3, 5, 7] \\
Maximum Kernel Size ($k_{\text{max}}$) & 7 \\
Depthwise Groups & 768 (channel-wise) \\
Pointwise CNN Output Channels & 128 \\
EECE Hidden Size & 64 \\
EECE Attention Dimension & 32 \\
Train/Dev/Test Split & 80\% / 10\% / 10\% \\
Augmentation Model & \texttt{T5-small} \\
Augmented Text Maximum Length & 512 \\
Cache Enabled for Data & Yes (CSV) \\
Cache Enabled for Labels & Yes (NPY) \\
\bottomrule
\end{tabular}
\label{tab:8}
\end{table}

\subsubsection{Baselines}
We list multiple baseline models and conduct ablations of the proposed model to comprehensively evaluate and analyze its performance. We rigorously tested our CISEA-MRFE model against the following representative LSTM, CNN, Lexicon-enhanced, BERT, GloVe, and other LLM-based SA models.

\begin{table}[htbp]
\centering
\scriptsize
\setlength{\tabcolsep}{2pt}
\renewcommand{\arraystretch}{1.35}
\caption{Comparison results for different model types across four datasets.}
\label{tab:my_label_2}
\resizebox{\linewidth}{!}{
\begin{tabular}{l l cc cc cc cc}
\toprule
\multirow{2}{*}{Type} & \multirow{2}{*}{Model} & \multicolumn{2}{c}{IMDb} & \multicolumn{2}{c}{Yelp} & \multicolumn{2}{c}{Twitter} & \multicolumn{2}{c}{Amazon} \\
& & Acc & M-F1 & Acc & M-F1 & Acc & M-F1 & Acc & M-F1 \\
\midrule
\multirow{7}{*}{BERT-based}
& AEN-BERT & 95.4 & 93.8 &  -   &  -   & 74.7 &  -   &  -   &  -   \\
& BERT & 95.7 & 94.2 & 71.4 & 68.9 & 67.5 & 64.3 & 67.8 & 64.7 \\
& BERT-SPC & 78.9 & 77.3 & 73.2 & 70.5 & 73.5 &  -   &  -   &  -   \\
& BERT-base+ITPT & 95.6 &  -   &  -   &  -   &  -   &  -   &  -   &  -   \\
& BERT-large+ITPT & 95.8 &  -   &  -   &  -   &  -   &  -   &  -   &  -   \\
& B-CNN & 93.3 & 93.0 &  -   &  -   &  -   &  -   & 91.4 & 90.2 \\
& B-MLCNN & 95.0 & 93.2 & 71.5 & 68.1 & 67.9 & 65.2 & 90.5 & 65.1 \\
\midrule
\multirow{3}{*}{Global Vector-based}
& AEN-GloVe &  -   &  -   &  -   &  -   & 72.8 &  -   &  -   &  -   \\
& MemNet &  -   &  -   &  -   &  -   & 69.3 &  -   &  -   &  -   \\
& GloVe & 85.3 &  -   &  -   &  -   &  -   &  -   &  -   &  -   \\
\midrule
\multirow{3}{*}{CNN-based}
& CNN-rand & 85.4 & 84.7 & 64.7 & 63.2 & 61.2 & 60.3 & 61.3 & 60.0 \\
& WDE-CNN & 89.6 &  -   & 67.1 & 65.3 &  -   &  -   &  -   &  -   \\
& CNN-multichannel & 88.5 & 86.8 & 65.4 & 64.1 & 64.3 & 63.5 & 64.8 & 64.1 \\
\midrule
\multirow{6}{*}{LSTM-based}
& C-LSTM & 86.7 &  -   & 63.4 &  -   &  -   &  -   &  -   &  -   \\
& AC-BiLSTM & 91.8 &  -   & 66.2 &  -   &  -   &  -   &  -   &  -   \\
& Tree-LSTM & 85.1 &  -   & 65.3 &  -   &  -   &  -   &  -   &  -   \\
& LSTM-GCA & 88.7 &  -   & 64.1 &  -   &  -   &  -   &  -   &  -   \\
& MBiLSTM & 93.6 &  -   &  -   &  -   &  -   &  -   &  -   &  -   \\
& AEC-LSTM & 96.3 &  -   & 73.5 &  -   &  -   &  -   &  -   &  -   \\
\midrule
\multirow{3}{*}{Lexicon-based}
& SAAT &  -   &  -   &  -   &  -   &  -   &  -   & 88.0 & 87.9 \\
& SAWE &  -   &  -   &  -   &  -   &  -   &  -   & 88.0 & 88.0 \\
& SAATWE &  -   &  -   &  -   &  -   &  -   &  -   & 88.1 & 88.0 \\
\midrule
\multirow{3}{*}{Prompt-based}
& Regular-PT & 92.4 &  -   &  -   &  -   & 91.8 &  -   &  -   &  -   \\
& P-Tuning & 91.0 &  -   &  -   &  -   &  -   &  -   & 93.0 &  -   \\
& SK-PT & 93.1 &  -   &  -   &  -   &  -   & 93.9 &  -   &  -   \\
\midrule
\multirow{4}{*}{Ours}
& CISEA-SADE & 88.5 & 88.3 & 75.1 & 74.2 & 93.5 & 93.4 & 90.3 & 89.7 \\
& CISEA-EECE & 90.1 & 89.9 & 75.8 & 75.2 & 93.7 & 93.9 & 89.8 & 88.5 \\
& CI-MRFE & \textbf{96.8} & \textbf{96.3} & 77.8 & 77.7 & 97.1 & 96.7 & 91.3 & 91.3 \\
& CISEA-MRFE & 91.3 & 90.5 & \textbf{78.0} & \textbf{77.6} & \textbf{97.8} & \textbf{97.1} & \textbf{93.2} & \textbf{92.5} \\
\bottomrule
\end{tabular}}
\end{table}

\begin{table}[htbp]
\centering
\scriptsize
\setlength{\tabcolsep}{3pt}
\renewcommand{\arraystretch}{1.35}
\caption{Results of our method combinations across five PLMs on four datasets. Accuracy and Macro-F1 (M-F1) show no statistically significant differences based on $t$-statistics and corresponding $p$-values.}
\label{tab:my_label_4}
\resizebox{\linewidth}{!}{
\begin{tabular}{l l cc cc cc cc}
\toprule
\multirow{2}{*}{Model} & \multirow{2}{*}{PLM} &
\multicolumn{2}{c}{IMDb} & \multicolumn{2}{c}{Yelp} & \multicolumn{2}{c}{Twitter} & \multicolumn{2}{c}{Amazon} \\
& & Acc & M-F1 & Acc & M-F1 & Acc & M-F1 & Acc & M-F1 \\
\midrule
\multirow{5}{*}{CISEA-MRFE}
& BERT       & 91.3 & 90.5 & 78.0 & 77.6 & 97.6 & 97.1 & 93.2 & 92.5 \\
& DistilBERT & 91.8 & 90.7 & 74.9 & 74.2 & 93.4 & 94.1 & 89.7 & 87.2 \\
& RoBERTa    & 93.3 & 92.5 & 75.7 & 75.1 & 95.1 & 95.8 & 90.7 & 89.2 \\
& BART       & 90.6 & 90.0 & 73.5 & 73.2 & 91.1 & 91.4 & 89.1 & 88.4 \\
& GPT-2      & 87.7 & 88.5 & 71.8 & 71.3 & 90.1 & 89.9 & 87.2 & 85.0 \\
\addlinespace[1pt]
& & $t$-stat. & 0.4443 & $t$-stat. & 0.3390 & $t$-stat. & -0.1050 & $t$-stat. & 0.9623 \\
& & $p$-val.  & 0.6686 & $p$-val.  & 0.7433 & $p$-val.  & 0.9190  & $p$-val.  & 0.3641 \\
\midrule
\multirow{5}{*}{CISEA}
& BERT       & 86.7 & 85.2 & 70.4 & 69.9 & 88.8 & 87.7 & 84.2 & 83.7 \\
& DistilBERT & 86.3 & 83.8 & 69.2 & 68.1 & 86.8 & 85.2 & 83.1 & 82.5 \\
& RoBERTa    & 86.7 & 84.5 & 70.4 & 69.9 & 87.8 & 86.2 & 83.6 & 83.1 \\
& BART       & 84.3 & 83.5 & 68.4 & 67.7 & 85.5 & 84.9 & 82.4 & 81.7 \\
& GPT-2      & 83.5 & 82.7 & 66.1 & 64.7 & 82.5 & 81.2 & 79.8 & 78.1 \\
\addlinespace[1pt]
& & $t$-stat. & 1.9645 & $t$-stat. & 0.6761 & $t$-stat. & 0.8085 & $t$-stat. & 0.6408 \\
& & $p$-val.  & 0.0851 & $p$-val.  & 0.5180 & $p$-val.  & 0.4422 & $p$-val.  & 0.5396 \\
\midrule
\multirow{5}{*}{Exclusive-PLM}
& BERT       & 82.2 & 81.2 & 66.9 & 66.3 & 82.5 & 81.7 & 78.5 & 78.2 \\
& DistilBERT & 80.9 & 80.4 & 64.6 & 65.1 & 82.3 & 81.3 & 79.5 & 78.7 \\
& RoBERTa    & 82.7 & 82.9 & 67.4 & 67.5 & 86.5 & 85.9 & 83.2 & 84.1 \\
& BART       & 76.3 & 76.6 & 60.2 & 59.7 & 79.3 & 79.1 & 76.8 & 75.4 \\
& GPT-2      & 80.7 & 80.9 & 65.1 & 64.9 & 84.3 & 84.5 & 80.6 & 79.5 \\
\addlinespace[1pt]
& & $t$-stat. & 0.1042 & $t$-stat. & 0.0759 & $t$-stat. & 0.2829 & $t$-stat. & 0.3049 \\
& & $p$-val.  & 0.9196 & $p$-val.  & 0.9419 & $p$-val.  & 0.7844 & $p$-val.  & 0.7682 \\
\bottomrule
\end{tabular}}
\end{table}

\begin{table}[htbp]
\centering
\scriptsize
\setlength{\tabcolsep}{2pt}
\renewcommand{\arraystretch}{1.05}
\caption{Ablation study on key components of CISEA-MRFE across four datasets.}
\label{tab:my_label_5}
\resizebox{\linewidth}{!}{
\begin{tabular}{l cc cc cc cc}
\toprule
\multirow{2}{*}{Model} &
\multicolumn{2}{c}{IMDb} &
\multicolumn{2}{c}{Yelp} &
\multicolumn{2}{c}{Twitter} &
\multicolumn{2}{c}{Amazon} \\
& Acc & M-F1 & Acc & M-F1 & Acc & M-F1 & Acc & M-F1 \\
\midrule
CISEA-MRFE                & 91.3 & 90.5 & 78.0 & 77.6 & 97.8 & 97.1 & 93.2 & 92.5 \\
w/o CI                    & 86.8 & 86.0 & 70.8 & 69.9 & 89.7 & 89.2 & 87.9 & 86.0 \\
w/o SEA                   & 96.8 & 96.3 & 77.8 & 77.5 & 97.1 & 96.7 & 91.3 & 91.3 \\
w/o SADE                  & 89.3 & 88.7 & 76.5 & 75.8 & 95.4 & 94.7 & 90.7 & 90.4 \\
w/o EECE                  & 89.1 & 88.4 & 76.2 & 77.1 & 95.1 & 94.1 & 90.2 & 89.9 \\
SADE w/ Single kernel (k = 3) & 89.5 & 89.2 & 76.8 & 75.9 & 95.8 & 95.1 & 91.3 & 90.8 \\
SADE w/ [1, 3, 5]         & 91.0 & 90.8 & 77.9 & 77.6 & 97.4 & 96.9 & 92.8 & 91.8 \\
\bottomrule
\end{tabular}}
\end{table}

\begin{itemize}
    \item Tree-LSTM \citep{KRAUS201965}: This model employs a complete discourse tree using a tensor-based tree-structured deep neural network.
    \item CNN-multichannel \citep{DBLP:conf/semeval/ZhangWZZ17}: This model employs a series of filters with varying window widths to extract a sequence of high-level textual features. 
    \item MBiLSTM \citep{PIMPALKAR2022117581}: This model employs a CNN layer and then integrating these elements into the multiple BiLSTM (MBiLSTM) model.
       \item AEN-Glove\citep{tang-etal-2016-effective}: This model employs an Attentional Encoder Network that skips recurrence by utilizing glove embedding.
         \item AEC-LSTM \citep{DBLP:journals/tnn/HuangLYZZQ22}: This model employs an attention mechanism with emotional recognition in convolutional LSTM  for sentiment analysis.
         \item AEN-BERT \citep{tang-etal-2016-effective}: This model employs an Attentional Encoder Network to describe the relationship amongst context and target using bert.
         \item BERT-SPC \citep{tang-etal-2016-effective}: This model employs filling out the basic BERT model with a specific sequence for the sentence pair classification task.
         \item BERT-base + ITPT \citep{DBLP:conf/cncl/SunQXH19}: Fine-tuned BERT-base model with Intermediate Task Prompt Tuning (ITPT) for domain adaptation.
         \item Glove \citep{DBLP:conf/emnlp/PenningtonSM14}: Employs unsupervised learning of word representations using word co-occurrence statistics.
         \item MemNet \citep{DBLP:journals/corr/abs-1708-02209}: A memory network-based model that dynamically captures relevant parts of the context.
         \item CNN-rand \citep{DBLP:conf/emnlp/Kim14}: A CNN model initialized with random word embeddings.
         \item WDE-CNN \citep{DBLP:journals/tkde/ZhaoGCHCWW18}: A word embedding distribution-enhanced CNN for aspect sentiment classification.
         \item C-LSTM \citep{DBLP:journals/corr/ZhouSLL15b}: Combines convolutional layers and LSTM to capture both local and long-distance dependencies.
         \item AC-BiLSTM \citep{DBLP:journals/ijon/LiuG19}: Attention-enhanced BiLSTM architecture for sentiment-specific feature extraction.    
        \item B-MLCNN \citep{ATANDOH2023101578} This model uses a bidirectional encoder representation from transformers with multiple channel convolutional neural network.
         \item SAAT \cite{10.1145/3357384.3357973}: Utilizes sentiment-aware attention for sentiment analysis.
         \item SAWE \cite{10.1145/3357384.3357973}: Utilizes sentiment-aware embedding.
          \item SAATWE \cite{10.1145/3357384.3357973}: Combines both sentiment-aware attention and embedding.
         \item Self Knowledgeable Prompt (SK)-PT:\citep{DBLP:journals/eswa/ZhuWMLQYW24}: The model design Soft Knowledgeable Prompt-tuning for concise text classification
         \item PT-Tuning \citep{liu2023gpt}: This model designs manually generated templates with trainable variables embedded in the embedded input. 
         \item RP-T\citep{schick2020exploiting}: This model designs customized templates and category names, with an unlabeled dataset annotated using an ensemble model.
         \item BERT \citep{devlin-etal-2019-bert}: BERT employs language models with masks to help with deep bidirectional representations that have been pre-trained.  
         \item BART \citep{lewis2019bart}: BART employs a pre-training model that integrates the advantages of two prominent decoders: the left-to-right GPT and the bidirectional BERT.
         \item RoBERTa \citep{DBLP:journals/corr/abs-1907-11692}: RoBERTa improves BERT optimization by eliminating the Next Sentence Prediction (NSP) task during pre-training, expanding the training dataset.

          \item DistilBERT \citep{sanh2019distilbert}: DistilBERT employs a technique for pre-training a small general-purpose language representation model that can be improved to perform well on a variety of tasks.

           \item GPT-2 \citep{radford2019language}: GPT-2 employs domain-specific high-quality data sets as input; this updated OpenAI GPT model is powered by a multi-layer unidirectional Transformer decoder.

\end{itemize}

\subsection{Experimental Results}
\subsubsection{Overall Performance (RQ1)}
Table~\ref{tab:my_label_2} reports the accuracy and macro-F1 scores across four benchmark datasets. The CISEA-MRFE model consistently outperforms competitive baselines, demonstrating superior flexibility and adaptability in capturing the subtleties of textual language.

Instruction-based and BERT-based models (Instruction, Prompt, BERT) show improved document- and sentence-level context understanding, effectively guiding model attention. On the IMDb dataset, CISEA-MRFE significantly outperforms models such as AEC-LSTM, BERT-large+ITPT, BERT-base+ITPT, B-MLCNN, Regular Prompts, P-Tuning, AC-BiLSTM, WDE-CNN, and BERT-SPC, with margins of: 0.3\% over P-Tuning, 1.1\% over WDE-CNN, and 12.4\% over BERT-SPC. In contrast, models like Tree-LSTM, GloVe, and BERT-SPC achieved the lowest accuracy.

On the Yelp dataset, BERT, B-MLCNN, and AEC-LSTM performed well, but the proposed model surpassed them with margins of 6.6\% over BERT and 20.8\% over BERT-SPC. CNN-based models (CNN-rand, WDE-CNN, CNN-multichannel) underperformed, likely due to their limitations in modeling long-range dependencies.

Despite slightly lower performance on Yelp possibly due to increased class cardinality and semantic overlap CISEA-MRFE still delivered robust results. On the Twitter dataset, CISEA-MRFE achieved significant accuracy gains: 23.1\% over AEN-BERT, 30.3\% over BERT, 27.3\% over BERT-SPC, 6.0\% over Regular-PT, and 25\% over AEN-GloVe.

On the Amazon dataset, our model outperformed several state-of-the-art baselines including BERT, B-CNN, SAAT, SAWE, SATAWE, B-MLCNN, and P-Tuning by margins of 28.5\%, 1.4\%, 5.24\%, 5.19\%, 5.14\%, 2.7\%, and 0.2\%, respectively. These results affirm the effectiveness and generalizability of CISEA-MRFE across domains.

Overall, CISEA‑MRFE demonstrates robust performance across datasets of differing length and complexity, confirming its adaptability and domain-general applicability.

\subsubsection{Ablation Studies (RQ2)}
To systematically investigate the contributions of individual components within the proposed CISEA-MRFE architecture, we conduct a comprehensive ablation study. This evaluation not only examines the effectiveness of each module but also assesses the impact of CISEA-MRFE when integrated with PLMs. As summarized in Table~\ref{tab:my_label_4}, we compare three primary configurations: (i) the full CISEA-MRFE model, (ii) a variant excluding the MRFE module (i.e., CISEA), and (iii) a PLM-only baseline without both CI and MRFE. These comparisons underscore the cumulative benefits of domain-specific guidance, semantic augmentation, and multi-refined representation learning.
From Table~\ref{tab:my_label_4}, CISEA-MRFE consistently achieves superior accuracy and macro-F1 across all datasets and PLMs, demonstrating strong cross-domain and cross-model generalization. Even when slight performance drops occur with certain PLMs, CISEA-MRFE remains competitive, reinforcing its robustness in modeling sentiment-relevant information.

To further dissect modular contributions, we perform targeted ablations as shown in Table~\ref{tab:my_label_5}, evaluating: (a) w/o CI, removal of contextual instruction; (b) w/o SEA, exclusion of Semantic Enhancement Augmentation; (c) w/o SADE, replacing the Scale-Adaptive Depthwise Encoder with a standard 1D convolutional layer; (d) w/o EECE, removing the Emotion Evaluator Context Encoder; (e) SADE (single kernel), using a fixed kernel size \( k=3 \); and (f) SADE (multi-kernel), using the default multi-kernel setting \( k \in \{1, 3, 5\} \). These variants enable fine-grained analysis of how local, global, and emotional representations interact to enhance sentiment classification.

Table~\ref{tab:my_label_5} reveals that the removal of the CI module results in the most pronounced performance degradation, with accuracy losses of 8.1\% on Twitter and 7.2\% on Yelp. This highlights the critical role of domain-guided disambiguation in handling noisy, informal, or context-sensitive sentiment cues.

Excluding SEA leads to performance drops of 0.2\% (Yelp), 0.7\% (Twitter), and 0.9\% (Amazon), suggesting that semantic augmentation enhances generalization through paraphrastic variation. The relatively minor impact on IMDb implies SEA’s particular utility in shorter or more informal text settings.

Removing SADE results in accuracy declines of up to 2.5\% (Yelp) and 2.4\% (Twitter), affirming the effectiveness of depth-wise convolutions for capturing sentiment-bearing n-gram structures. Multi-kernel configurations provide richer local features, particularly beneficial in heterogeneous or multi-aspect reviews.

The absence of EECE yields moderate but consistent degradation of 1.9\% on IMDb and 1.7\% on Amazon emphasizing the value of global context modeling and emotion-aware attention in resolving subtle or distributed sentiment expressions.

Overall, the complete CISEA-MRFE model outperforms all ablated versions, achieving absolute gains of 4.6\% (IMDb), 6.5\% (Yelp), 30.3\% (Twitter), and 4.1\% (Amazon) over the weakest configurations. These findings validate the synergistic effect of integrating contextual instruction, semantic enrichment, hierarchical local-global encoding, and affective reasoning.

In summary, each module contributes uniquely to robustness and generalization. Their combined effect enables CISEA-MRFE to adapt across diverse domains and text structures with varying sentiment expression styles, thereby advancing the frontier of modular sentiment classification.

\subsubsection{Computational Efficiency Analysis and Comparison (RQ3)}

\paragraph{Computational Efficiency Analysis}
We benchmark CISEA-MRFE and its ablated variants (without SADE or EECE) across four datasets to evaluate trade-offs between representational power and computational efficiency (Table~\ref{tab:6}).

The full model consistently achieves the highest accuracy, demonstrating the synergistic roles of SADE and EECE. SADE contributes to both local compositional feature capture and early-stage feature compression, reducing inference overhead despite its multi-scale design. In contrast, EECE enhances emotional reasoning and global context modeling but introduces higher latency due to its recurrent structure. Removing SADE unexpectedly increases inference time and parameter count (109.6M to 120.3M), underscoring the efficiency of depth-wise convolutional encoding. Excluding EECE reduces model size and latency but results in moderate performance degradation, reflecting a trade-off between emotional granularity and speed.

Overall, CISEA-MRFE strikes a favorable balance between performance and efficiency, achieving inference latencies between 1.10ms and 2.62ms per sample, while maintaining sentiment granularity. These findings advocate for modularly informed architectural design in sentiment classification where components like SADE offer computational compactness, and EECE ensures semantic-emotional depth. Such insights are particularly valuable when optimizing for resource-constrained or latency-sensitive environments.

\begin{table}[htbp]
\centering
\scriptsize
\setlength{\tabcolsep}{2pt}
\renewcommand{\arraystretch}{1.05}
\caption{Efficiency comparison of CISEA-MRFE and its ablated variants across four benchmark datasets. Results are reported as mean ± standard deviation.}
\label{tab:6}
\resizebox{\linewidth}{!}{
\begin{tabular}{l l c c c c}
\toprule
Configuration & Dataset & Accuracy (\%) & Inference Time (s) & Runtime/Sample (ms) & Params (M) \\
\midrule
\multirow{4}{*}{Full Model} 
& IMDb    & 91.3 ± 0.2   & 250.65  & 2.35 ± 0.04   & 109.6 \\
& Yelp    & 78.0 ± 0.3   & 185.93  & 1.50 ± 0.02   & 109.6 \\
& Twitter & 97.8 ± 0.1   & 631.63  & 2.62 ± 0.03   & 109.6 \\
& Amazon  & 93.2 ± 0.2   & 24.57   & 1.10 ± 0.01   & 109.6 \\
\midrule
\multirow{4}{*}{w/o EECE} 
& IMDb    & 89.1 ± 0.2   & 180.21  & 1.72 ± 0.03   & 91.4  \\
& Yelp    & 76.2 ± 0.4   & 134.12  & 1.08 ± 0.02   & 91.4  \\
& Twitter & 95.1 ± 0.3   & 512.44  & 2.13 ± 0.03   & 91.4  \\
& Amazon  & 90.2 ± 0.3   & 18.93   & 0.85 ± 0.01   & 91.4  \\
\midrule
\multirow{4}{*}{w/o SADE} 
& IMDb    & 89.3 ± 0.3   & 271.09  & 2.50 ± 0.05   & 120.3 \\
& Yelp    & 76.5 ± 0.3   & 204.87  & 1.65 ± 0.03   & 120.3 \\
& Twitter & 95.4 ± 0.2   & 712.18  & 2.95 ± 0.04   & 120.3 \\
& Amazon  & 90.7 ± 0.3   & 26.91   & 1.20 ± 0.02   & 120.3 \\
\bottomrule
\end{tabular}}
\end{table}

\paragraph{Computational Efficiency Comparison}

To evaluate the computational efficiency of the proposed CISEA-MRFE architecture, we benchmarked it against five representative baselines, including BERT, AEC-LSTM, BERT-CNN, BERT-SPC, and AEN-BERT on the IMDb dataset. 
As shown in Table~\ref{tab:efficiency_metrics}, CISEA-MRFE maintains a competitive parameter size (109.6M) while exhibiting the highest FLOPs (27.6G), primarily due to the inclusion of enriched multi-refinement modules that enhance representational capacity~\citep{he2016deep, howard2017mobilenets, dai2021coatnet}.

Notably, it attains the lowest inference latency (1.89 ms/sample) among all models and a moderate training time of 3.7 minutes per epoch. This apparent trade-off higher FLOPs accompanied by faster inference is attributed to the architecture's use of depth-wise separable convolutions and channel-wise operations, which reduce memory access costs and enhance GPU parallelism. These design choices are consistent with efficient modeling principles observed in prior work such as MobileNet \citep{howard2017mobilenets} and MobileNetV2 \citep{sandler2018mobilenetv2}, where the separation of spatial and channel dimensions leads to both performance gains and runtime savings. In contrast, baseline models like AEN-BERT and BERT-SPC exhibit higher inference latency (10.5 ms and 9.8 ms/sample, respectively) despite comparable or lower FLOPs and parameter counts. This suggests less favorable compute-to-latency optimization in their architectural design.

In summary, the proposed CISEA-MRFE achieves a favorable trade-off between computational complexity and real-time applicability, demonstrating that architectural innovation rather than parameter count alone plays a decisive role in system-level efficiency for sentiment analysis tasks.
These results affirm CISEA-MRFE’s viability for real-world deployment in time-sensitive sentiment analysis scenarios, especially where both performance and efficiency are critical. 

\begin{table}[htbp]
\centering
\scriptsize
\setlength{\tabcolsep}{3pt}
\renewcommand{\arraystretch}{1.05}
\caption{Comparison of computational efficiency between the proposed CISEA-MRFE and five strong baselines on the IMDb dataset. Reported metrics include parameter count, floating-point operations per forward pass (FLOPs), average inference latency per sample, and training time per epoch. All models are evaluated under consistent conditions using fixed-length tokenized inputs across 1000 test instances.}
\label{tab:efficiency_metrics}
\resizebox{\linewidth}{!}{
\begin{tabular}{l c c c c}
\toprule
Model & Params (M) & FLOPs (G) & Inference (ms/sample) & Train Time (min/epoch) \\
\midrule
BERT        & 109.0 & 21.8 & 9.6  & 2.5 \\
AEC-LSTM    & 14.5  & 1.9  & 7.1  & 1.8 \\
BERT-CNN    & 111.3 & 22.6 & 9.2  & 2.6 \\
BERT-SPC    & 109.5 & 22.4 & 9.8  & 2.6 \\
AEN-BERT    & 112.1 & 24.0 & 10.5 & 2.9 \\
\textbf{CISEA-MRFE} & \textbf{109.6} & \textbf{27.6} & \textbf{1.89} & \textbf{3.7} \\
\bottomrule
\end{tabular}}
\end{table}

\subsubsection{Parametric Sensitivity (RQ5)}
\begin{figure}[htbp]
    \centering
    \includegraphics[width=\linewidth]{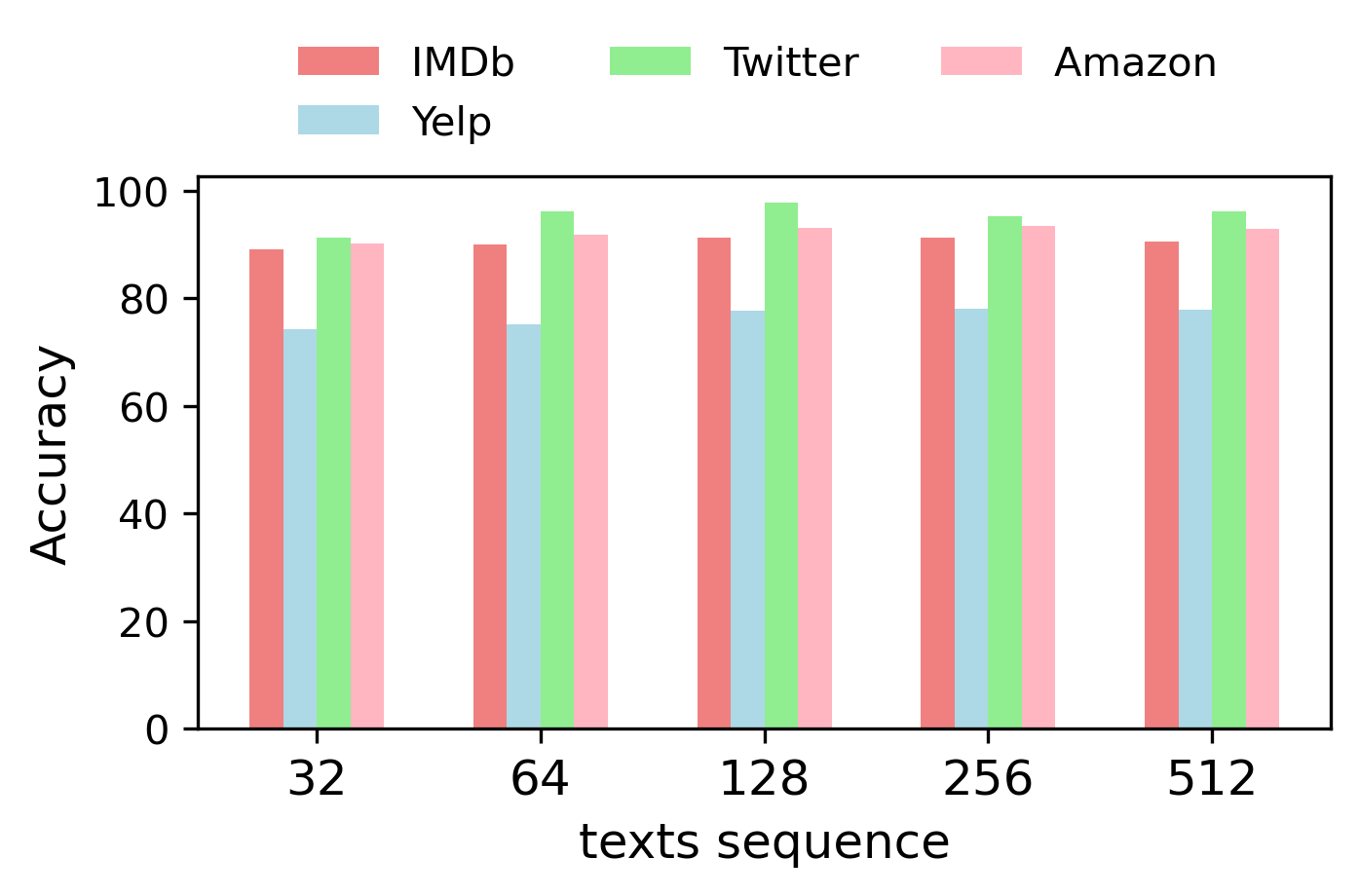}
    \caption{Effect of input sequence length on classification accuracy across datasets. EECE benefits from longer sequences in datasets like IMDb, while SADE performs optimally at moderate lengths for shorter, noisier texts such as Twitter and Yelp.}
    \label{Figure_3}
\end{figure}

\paragraph{Impact of Input Sequence Length}
As shown in Figure~\ref{Figure_3}, performance trends vary across datasets with respect to input length. On the IMDb dataset comprising long-form narrative reviews, longer sequences improve performance due to the EECE's ability to model extended contextual dependencies. In contrast, Yelp, Twitter, and Amazon datasets achieve optimal accuracy at moderate sequence lengths. Short sequences result in inadequate contextual cues, while excessively long inputs introduce noise and semantic dilution. These observations validate the design of EECE as a recurrent module that benefits from informative but focused textual context, and the need to balance context size with noise sensitivity in diverse domains.

\begin{figure}[htbp]
    \centering
    \includegraphics[width=0.95\linewidth]{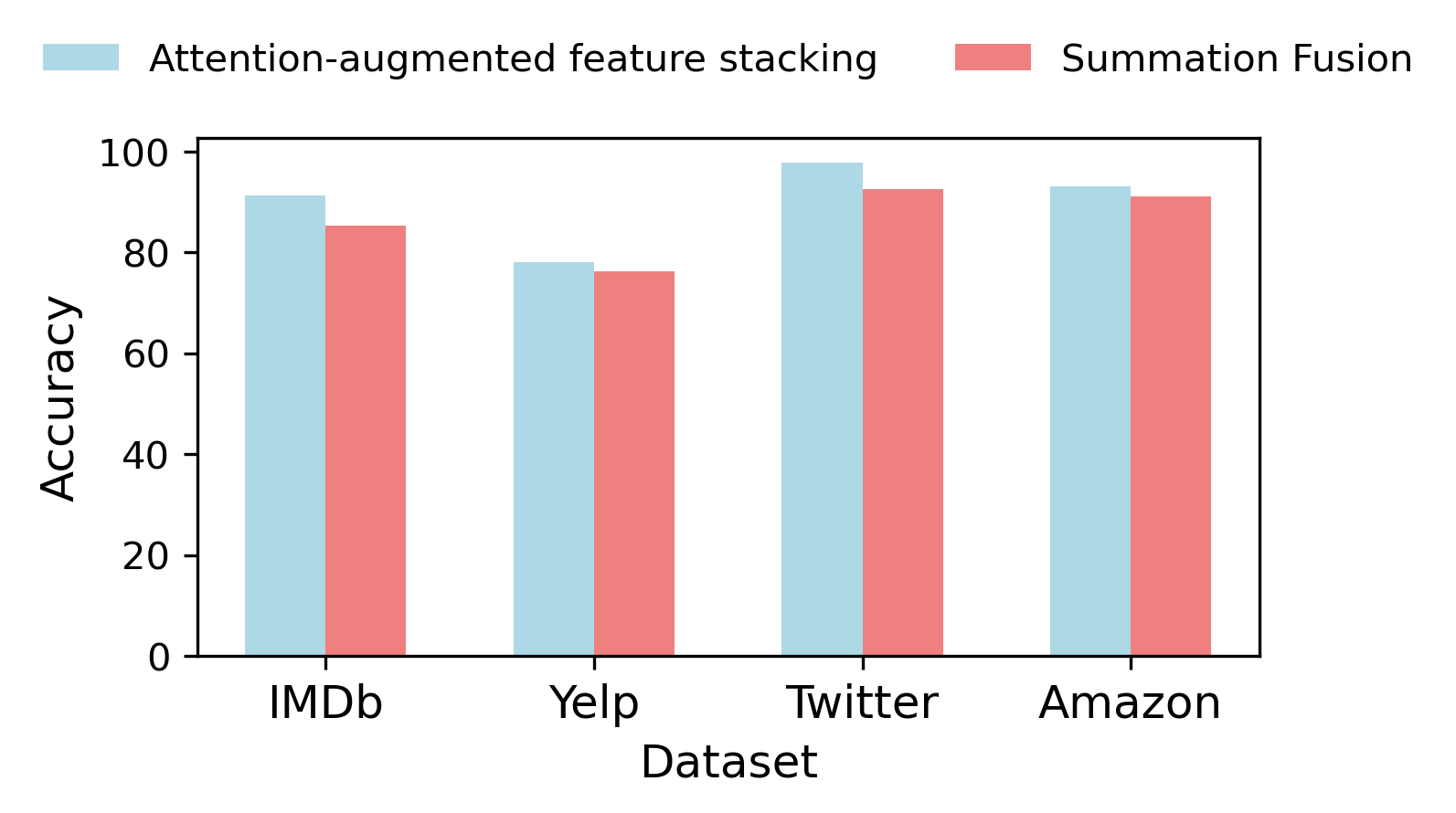}
    \caption{Comparison of feature fusion strategies between SADE and EECE outputs. Attention-based fusion yields superior performance across datasets by adaptively weighting local and global-emotional features.}
    \label{Figure_4}
\end{figure}

\paragraph{Impact of Fusion Mechanism}
Figure~\ref{Figure_4} presents a comparative evaluation of fusion strategies for integrating outputs from the SADE and EECE modules. Attention-based feature fusion consistently outperforms alternatives, dynamically assigning importance weights to local (SADE) and global-emotional (EECE) representations. This adaptive fusion aligns with the heterogeneous nature of sentiment features, where either local patterns (e.g., negation phrases) or global narrative (e.g., tone progression) may dominate. On short-form datasets like Twitter, a simple summation fusion yields competitive results due to concentrated sentiment expressions and reduced need for deep compositional reasoning. These results highlight the value of emotion-guided fusion and justify the inclusion of flexible integration mechanisms in multi-stream sentiment architectures.

\begin{figure}[htbp]
    \centering
    \includegraphics[width=0.95\linewidth]{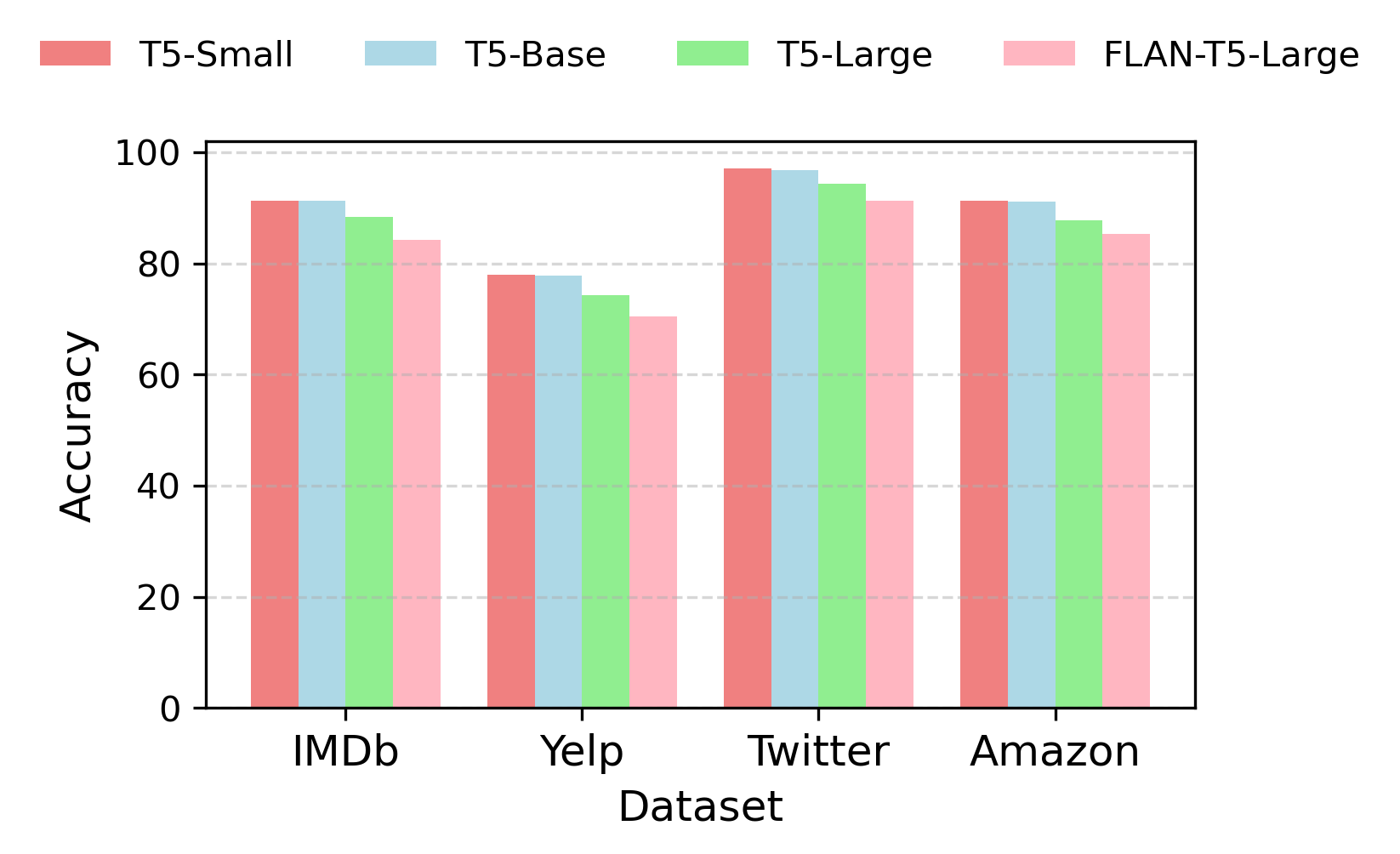}
    \caption{Performance of different language models in the text generation pipeline for data augmentation.}
    \label{Figure_6}
\end{figure}

\paragraph{Impact of Text-to-Text Generation in Semantic Data Augmentation}

To evaluate the contribution of generative data augmentation, we conducted an empirical study leveraging text-to-text generation models within the BERT-based training pipeline. Specifically, we assessed multiple T5 variants—namely T5-small, T5-base, T5-large, and FLAN-T5-large—to dynamically generate paraphrased sequences that enhance input diversity during training.

As illustrated in Figure~\ref{Figure_6}, all generative variants contributed positively to classification performance by enriching the training distribution with semantically consistent yet lexically diverse examples. This augmentation strategy introduces controlled syntactic variations, enabling the model to better generalize to unseen instances and mitigating overfitting in low-resource or noisy domains.

Among the evaluated models, T5-small demonstrated the most favorable trade-off between performance and efficiency, achieving the highest accuracy and macro-F1 while incurring the lowest computational overhead. Notably, T5-small reduced average training time by approximately 38\% relative to larger models, while delivering a 2.3\% average improvement in accuracy across datasets. These findings suggest that lightweight generative models can yield substantial gains in robustness and predictive quality without the latency and resource demands of their larger counterparts.

Overall, the results underscore the value of efficient and semantically faithful text augmentation for enhancing generalization in sentiment classification tasks, thereby reinforcing the practical utility of the CISEA-MRFE architecture in resource-constrained environments.

\paragraph{Effect of Channel-Wise Depth-Wise Convolutional Kernels on Sentiment Classification Accuracy}

To systematically assess the impact of receptive field diversity within the SADE module, we conducted a comprehensive setting involving various kernel configurations across four representative sentiment classification datasets: IMDb, Yelp, Amazon, and Twitter. Each configuration employed distinct combinations of kernel sizes \( k \in \{1,3,5,7\} \), where the maximum kernel value approximates the effective receptive field for local feature extraction. This setup enables the exploration of both narrow and broad contextual scopes through depth-wise separable convolutions.

\begin{figure}[htbp]
    \centering
    \begin{subfigure}[b]{0.23\linewidth}
        \centering
        \includegraphics[width=\linewidth]{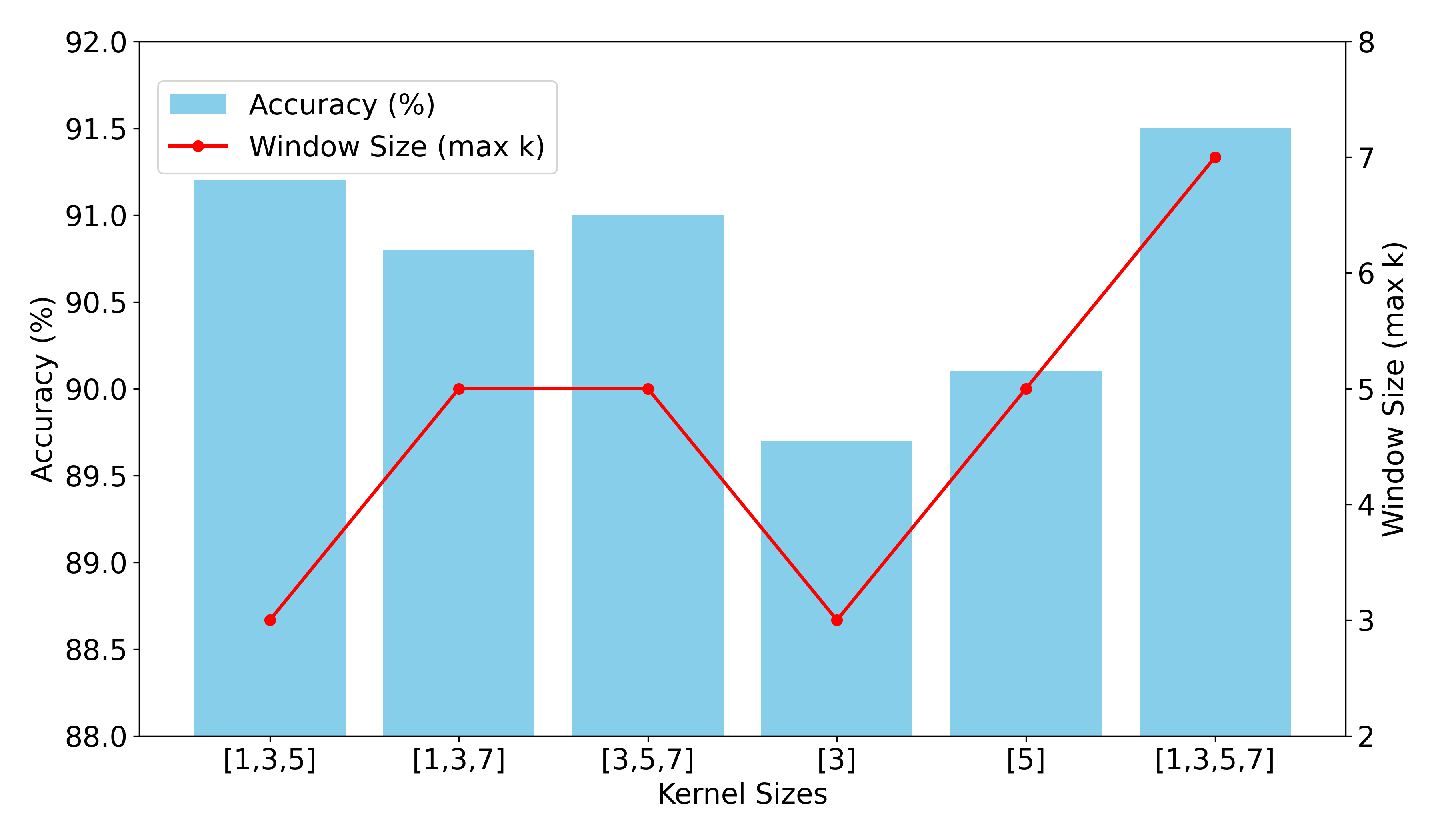}
        \caption{IMDb}
        \label{fig:sub7}
    \end{subfigure}\hfill
    \begin{subfigure}[b]{0.23\linewidth}
        \centering
        \includegraphics[width=\linewidth]{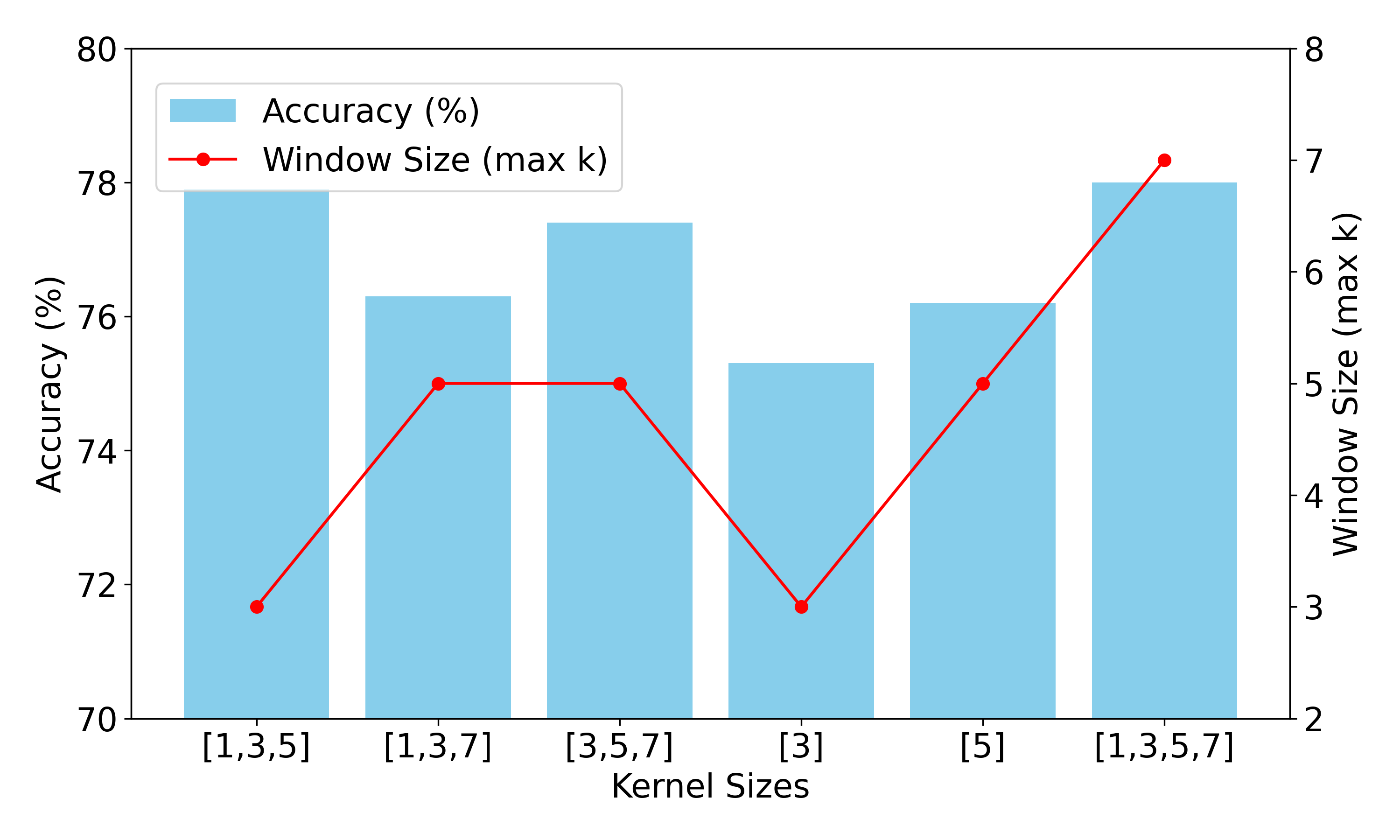}
        \caption{Yelp}
        \label{fig:sub8}
    \end{subfigure}\hfill
    \begin{subfigure}[b]{0.23\linewidth}
        \centering
        \includegraphics[width=\linewidth]{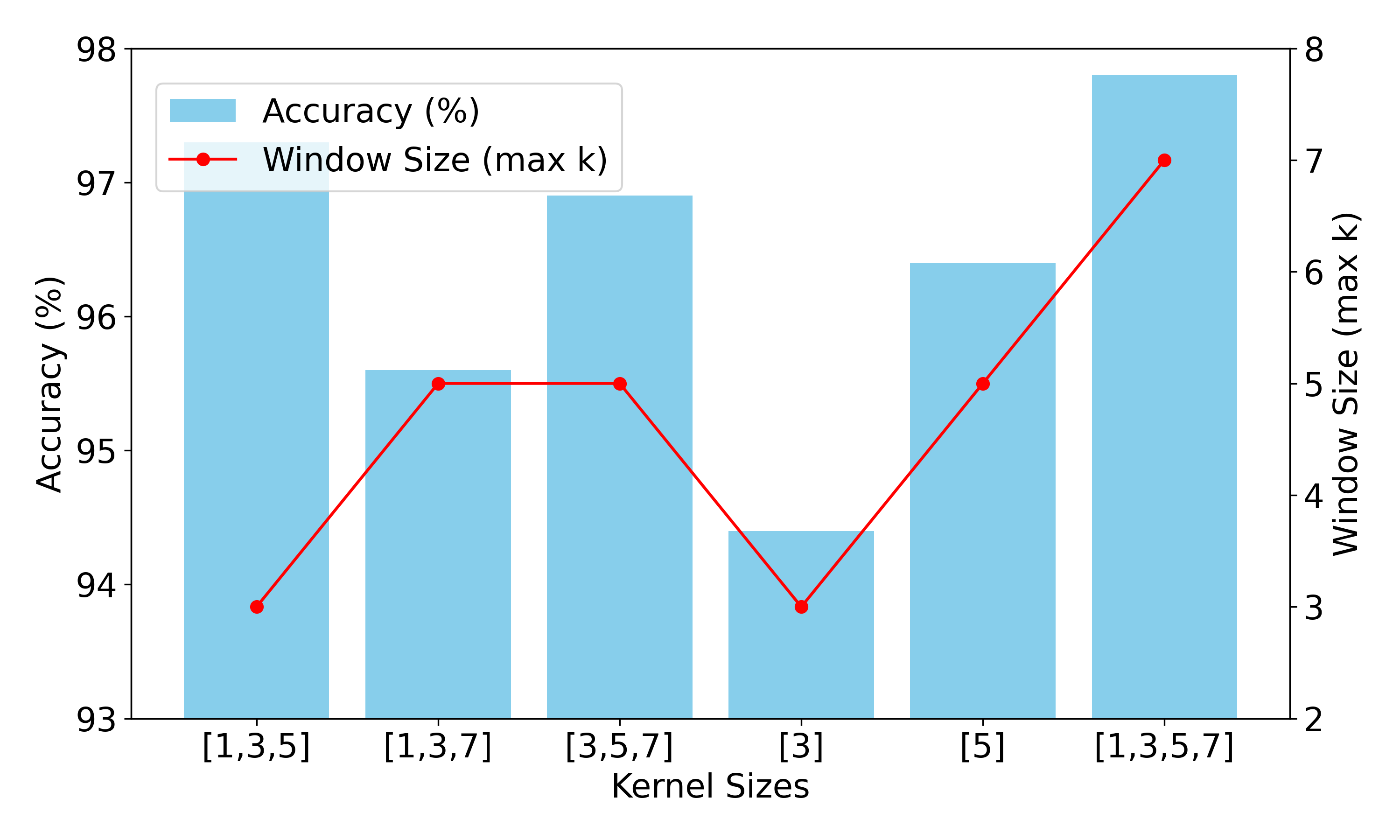}
        \caption{Twitter}
        \label{fig:sub12}
    \end{subfigure}\hfill
    \begin{subfigure}[b]{0.23\linewidth}
        \centering
        \includegraphics[width=\linewidth]{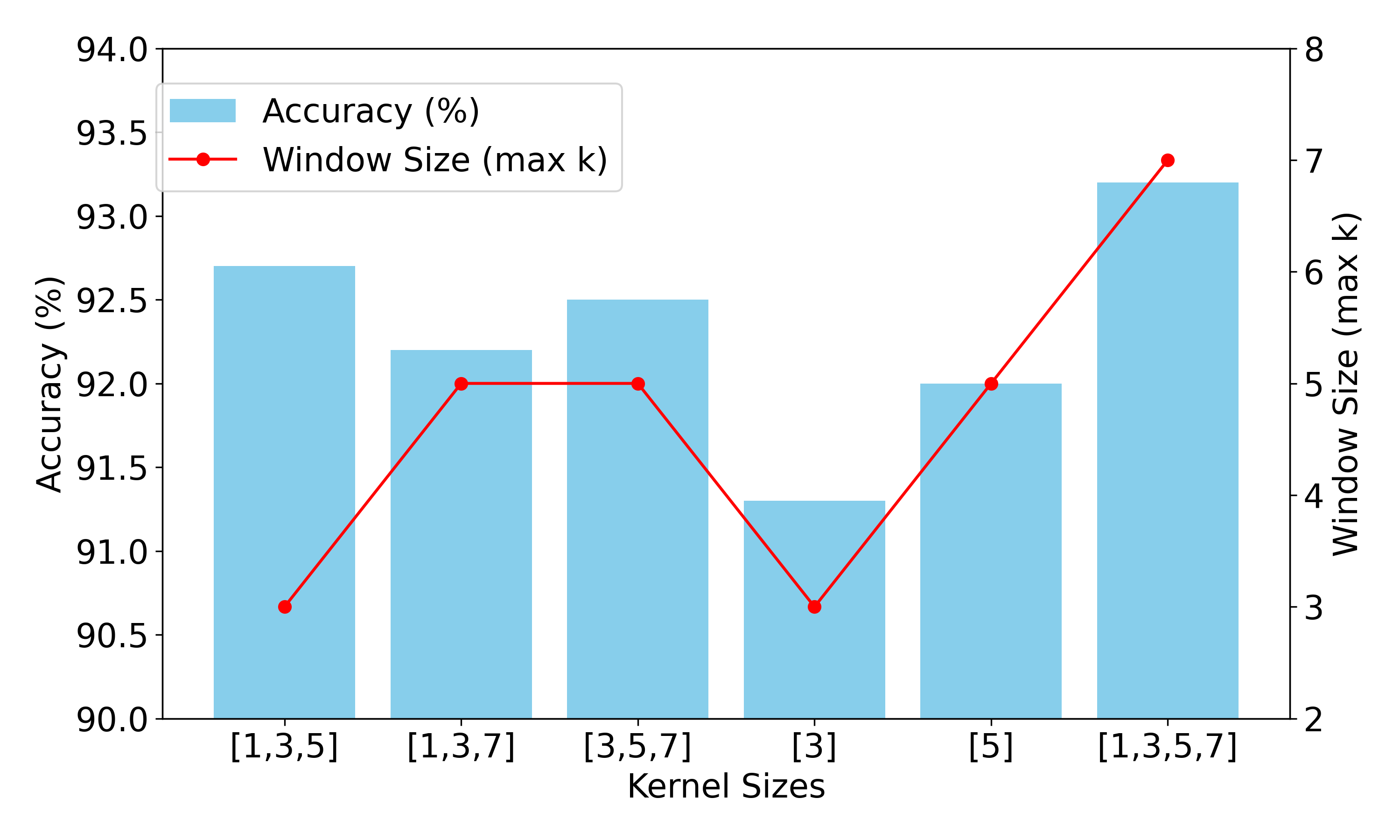}
        \caption{Amazon}
        \label{fig:sub13}
    \end{subfigure}
    \caption{Comparative performance analysis of various multi-kernel configurations in the SADE module across four benchmark datasets. The bar plot illustrates classification accuracy achieved by each kernel combination, while the overlaid line plot indicates the effective window size (maximum kernel span). The results indicate that integrating broader and more diverse receptive fields enhances local feature extraction, improving model performance across tasks.}
    \label{fig:5}
\end{figure}

As shown in Figure~\ref{fig:5}, models incorporating a broader and more diverse kernel spectrum particularly the configuration, \([1,3,5,7]\) consistently outperformed their single-scale counterparts across all datasets. For example, accuracy on the IMDb dataset increased from 75.2\% with a single kernel size \([3]\) to 78.0\% with the multi-scale configuration \([1,3,5,7]\). Similar trends were observed on the Twitter and Amazon datasets, where richer receptive hierarchies enabled more effective modeling of syntactic and emotional cues, particularly within complex or context-rich reviews.

Configurations constrained to a single kernel size (e.g., \([3]\), \([5]\)) exhibited diminished performance, highlighting their limited capacity to capture heterogeneous sentiment expressions. Intermediate configurations such as \([1,3,5]\) and \([3,5,7]\) demonstrated moderate improvements, reinforcing the hypothesis that multi-granularity filtering enhances representational flexibility. However, marginal returns diminished beyond a certain scale, as indicated by the plateau in accuracy gains upon the inclusion of \(k=7\), likely due to increased feature redundancy or contextual over-smoothing.

Overall, the findings suggest that multi-scale depth-wise convolutional encoding enhances the expressiveness of sentiment representations by capturing varying levels of contextual granularity. Among the configurations evaluated, \([1,3,5,7]\) demonstrates the most balanced performance across datasets, offering a favorable trade-off between receptive field diversity and discriminative capacity. Accordingly, this configuration emerges as a practical choice for default kernel design within the sentiment classification architecture.

\begin{figure}[htbp]
    \centering
    \includegraphics[width=0.95\linewidth]{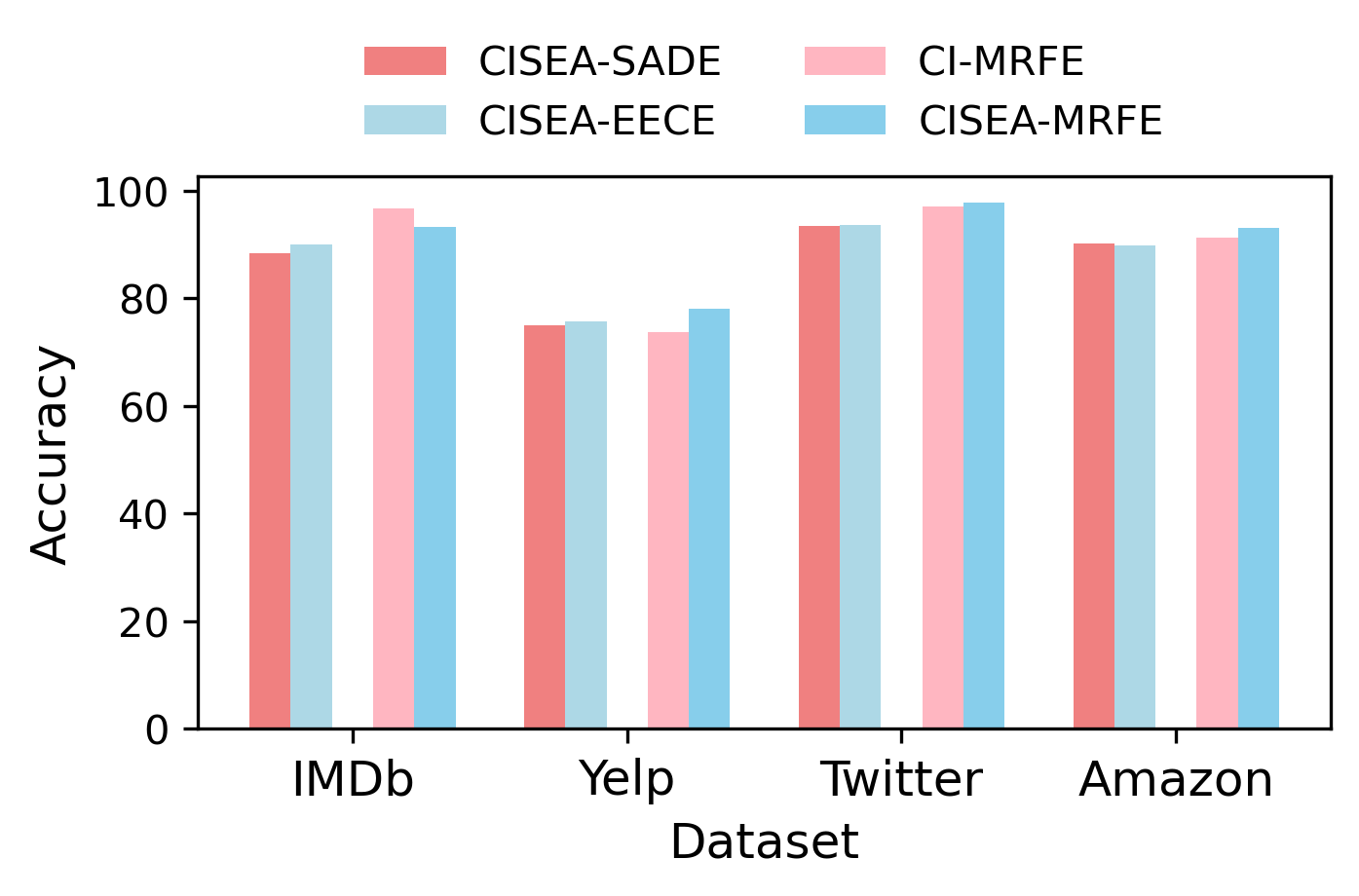}
    \caption{Performance of CISEA models across four benchmark datasets.}
    \label{Figure_5}
\end{figure}

\paragraph{Impact of Model Performance Across Datasets}
We conducted a comprehensive analysis of the proposed model and its baseline variants (CISEA-SADE, CISEA-EECE, CI-MRFE, and CISEA-MRFE) across four benchmark datasets (Figure~\ref{Figure_5}). 

The CISEA-SADE variant shows relatively lower performance on the IMDb dataset, likely due to its limited capacity to capture long-range dependencies in extended review texts. In contrast, it performs competitively on Yelp, Twitter, and Amazon, where shorter, capsule-like reviews align better with its local feature extraction design. CISEA-EECE consistently outperforms CISEA-SADE, with notable improvements on IMDb (1.6\% accuracy gain), highlighting the advantage of integrating global emotional context for modeling long-form narratives.

The CI-MRFE variant delivers marked improvements over CISEA-SADE across all datasets, particularly achieving a 4.9\% accuracy gain on IMDb, underscoring the value of multi-refined feature extraction for hierarchical representation learning. The full CISEA-MRFE model achieves the highest performance across all datasets, offering balanced gains in both accuracy and macro-F1. Notably, it delivers substantial improvements on IMDb and Amazon, while providing moderate yet consistent gains on Yelp and Twitter, confirming its robustness across diverse text lengths and domains.

These findings demonstrate the complementary roles of contextual instruction, semantic augmentation, and multi-refined feature extraction in enhancing sentiment classification, with CISEA-MRFE showing superior adaptability to varying domain and input characteristics.

\subsection{Discussion}
\label{sec:discussion}

This study introduces CISEA‑MRFE, a comprehensive sentiment classification framework that integrates CI, SEA, and an MRFE module composed of SADE and EECE. By combining contextual guidance, paraphrastic augmentation, and multi‑granular feature refinement, CISEA‑\newline MRFE is designed to address key challenges in sentiment analysis, including domain transferability, the modeling of nuanced emotional cues, and performance degradation under data imbalance.
Empirical evaluations across four diverse benchmark datasets, including IMDb, Yelp, Twitter, and Amazon, demonstrate that CISEA‑MRFE consistently outperforms competitive baselines, underscoring its ability to effectively capture both short‑ and long‑form sentiment cues. Importantly, the model maintains high performance across multiple pre‑trained language model backbones (BERT, RoBERTa, BART, DistilBERT, GPT‑2), confirming its robustness and architectural transferability. These findings highlight the value of the CI module in disambiguating sentiment polarity under noisy settings, while SEA contributes to mitigating bias toward dominant sentiment classes through controlled semantic augmentation.
Ablation analyses further reveal the critical role of the MRFE module, where SADE improves localized n‑gram sensitivity and EECE refines global emotional context. Removing these components results in significant performance degradation, confirming that multi‑scale feature refinement and emotion‑aware modeling are essential to the framework’s efficacy. Notably, even simplified CISEA‑MRFE variants surpass recent prompting‑based strategies, illustrating the superiority of explicit multi‑level feature integration over template‑only approaches.
In summary, CISEA‑MRFE unifies instruction‑guided learning, emotion‑aware encoding, and multi‑scale feature extraction, delivering a scalable and versatile solution for real‑world sentiment analysis.

\section{Conclusion}

This paper proposed CISEA-MRFE, a novel sentiment analysis framework that integrates Contextual Instruction (CI), Semantic Enhancement Augmentation (SEA), and a Multi-Refined Feature Extraction (MRFE) module comprising Scale-Adaptive Depthwise Encoding (SADE) and Emotion-Aware Context Encoding (EECE). By combining instruction-guided encoding, semantic augmentation, and multi-scale feature learning, the model effectively captures local, global, and affective cues critical for robust sentiment classification.

Extensive experiments across four benchmark datasets IMDb, Yelp, Twitter, and Amazon demonstrate that CISEA-MRFE consistently outperforms competitive baselines in both accuracy and macro-F1 metrics. Ablation studies further confirm the complementary contributions of CI, SEA, SADE, and EECE to sentiment understanding across short and long textual reviews.

Beyond accuracy, our analysis also highlights the model’s computational efficiency. Specifically, SADE contributes to early-stage compression with minimal latency overhead, while EECE adds emotional granularity with acceptable parameter cost. Runtime evaluations show that CISEA-MRFE achieves a favorable trade-off between performance and efficiency, maintaining inference times within practical bounds (1.10–2.62 ms/sample) and moderate parameter growth, making it suitable for real-world deployment in resource-constrained scenarios.

Despite its strengths, the framework relies on well-designed instruction templates, where misaligned guidance can reduce effectiveness. Also, the text-to-text data augmentation for long inputs in the framework may introduce noise and increase computational burden. Future work will focus on developing adaptive instruction generation mechanisms, exploring hierarchical augmentation strategies, and extending the framework to aspect-level and multilingual sentiment classification. These directions aim to further improve the model’s reasoning capacity, generalizability, and scalability across diverse sentiment analysis tasks. \newline
\newline
\newline


\end{document}